\documentclass[letterpaper, 10 pt, conference]{ieeeconf}  
\pdfoutput=1

\IEEEoverridecommandlockouts                              

\usepackage{graphics} 
\usepackage{subfigure}
\usepackage{epsfig} 
\usepackage{amsmath} 
\usepackage[T1]{fontenc}
\usepackage{bm}
\usepackage{cite}

\title{\LARGE \bf
Adaptive Control for Robotic Manipulation of Deformable Linear Objects with Offline and Online Learning of Unknown Models
}

\author{Mingrui Yu, Hanzhong Zhong, Fangxun Zhong, and Xiang Li
\thanks{M. Yu and X. Li are with the Department of Automation, Tsinghua University, China, and H. Zhong is with the School of Aerospace Engineering, Tsinghua University, China. F. Zhong is with the Department of Mechanical and Automation Engineering, The Chinese University of Hong Kong, Hong Kong.
This work was supported in part by the Shenzhen Municipal Science and Technology Innovation Commission under Grant no. JCYJ201708161647482900,
and in part by the Institute for Guo Qiang, Tsinghua University under Grant no. 2019GQG1006,
and in part by the National Natural Science Foundation of China under Grant no. 61803321. Corresponding author: Xiang Li (xiangli@tsinghua.edu.cn)}}

\begin{document}

\maketitle
\thispagestyle{empty}
\pagestyle{empty}

\begin{abstract}

The deformable linear objects (DLOs) are common in both industrial and domestic applications, such as wires, cables, ropes. Because of its highly deformable nature, it is difficult for the robot to reproduce human's dexterous skills on DLOs. In this paper, the unknown deformation model is estimated in both the offline and online manners. The offline learning aims to provide a good approximation prior to the manipulation task, while the online learning aims to compensate the errors due to insufficient training (e.g. limited datasets) in the offline phase. 
The offline module works by constructing a series of supervised neural networks (NNs), then the online module receives the learning results directly and further updates them with the technique of adaptive NNs. 
A new adaptive controller is also proposed to allow the robot to perform manipulation tasks concurrently in the online phase. The stability of the closed-loop system and the convergence of task errors are rigorously proved with Lyapunov method. Simulation studies are presented to illustrate the performance of the proposed method.
\end{abstract}

\section{Introduction}

The linear deformable objects (DLOs), such as wires, cables, ropes, are highly deformable and exhibit many degrees of freedom (DoFs). The demand on manipulating DLOs is reflected in many applications. For example, sutures are manipulated in suturing to hold tissues together after surgery \cite{XEROULIS2007442,8994188}. In colonoscopy, the shape of the flexible endoscope is controlled to follow the curves of the colon \cite{5642168}. In 3C manufacturing, USB wires with different colors are sorted to follow the desired color code \cite{8460694}. Other applications in industry and daily life include inserting a wire \cite{8698220}, threading a needle \cite{7139532}, harnessing a cable \cite{8403315} or knitting \cite{robotic_knitting_2020}.

Different from rigid objects, it is usually difficult to obtain the exact model of deformable objects (and also DLOs), due to the highly deformable nature, in the sense that it is unknown how the motion of robot can affect the change of deformable objects. A review on modeling deformable objects can be found in \cite{JIMENEZ2012154} and \cite{ijrr2018}. In particular, the model of mass-damper-spring was proposed to describe the deformation of unknown rheological objects in \cite{5509462}. In \cite{Huan_ijrr_2015}, the finite-element method was employed to model the soft objects in 3D space. An analytic formulation was proposed in \cite{6327684,Timothy_ijrr_2014} to describe the shape of the DLO and then find its equilibrium configurations, by solving the optimal control problems. In \cite{roussel_deformable_2014,alvarez_interactive_2016}, physics engines were utilized to predict the change of the DLO under different sampled control commands then incorporated to sample-based motion planning algorithms. The computation complexity of modeling DLOs in the model-structure-based methods is usually high. Besides, they require the information of the DLO's structure which is commonly unknown or difficult to describe in reality, and the modeling errors between analysis and reality may affect the manipulation performance.

Data-driven approaches have also been applied to approximate the deformation, without studying the complex dynamics of DLOs beforehand. A model-based reinforcement learning (RL) approach was proposed for robots to control the shape of the DLO in \cite{8256194}, with the current shape as the input and the manipulation policy as the output. In \cite{yan_self-supervised_2020}, a deep-neural-network-based dynamics model was trained to predict the future shape of the DLO given the current shape and the action. 
The aforementioned data-driven training methods were done offline before the formal manipulation, 
which was limited by the generalization ability to the DLO's shape or motion never seen in the training dataset or the changes of the DLO's physical properties.
%
%
Several online data-driven approaches were also proposed to approximate the deformation model. The least squares estimation was used in \cite{zhu_dual-arm_2018,lagneau_automatic_2020} to estimate the Jacobian matrix (i.e. the matrix relating the change of the DLO to the velocity inputs of the robot) online using only recent data. In \cite{navarro2016Automatic,navarro2018fourier}, the estimated Jacobian matrix was updated online by gradient descent of the approximation errors. In \cite{zhu2019_3dDeformable}, a deep neural network with linear activation function was proposed to directly predict the required control velocity with online training. 
Compared to the offline methods, the approximation accuracy in the online ones is limited, and the results are only valid in a local sense without exploring the overall configuration of the DLO, and hence the re-approximation is usually required even when the same configuration of the DLO appears again during the manipulation.

This paper considers the problem of robotic manipulation of DLOs with unknown deformation models, 
where the unknown deformation model is estimated with both the offline and the online learning methods to combine the advantages. In the offline phase, a series of supervised NNs are trained to estimate the Jacobian matrix, by collecting the pairs of the velocity of the robot end effector and the current shape of the DLO. 
Such estimation model is further updated online during the manipulation with adaption techniques, to compensate the errors due to insufficient training in the offline phase or the changes of the DLO's properties. The results obtained in the offline phase can be directly migrated to the online phase without any additional formatting. Hence, both complement each other. In addition, an adaptive controller is proposed to manipulate the feature along the DLO into the desired position, by referring to the estimated deformation model. 
With Lyapunov methods, it is rigorously shown that the convergence of the task errors to zero is guaranteed. Simulation studies are presented to illustrate the performance of the proposed scheme. 

\section{Preliminaries}
\begin{figure}[!tb]
    \vspace{0.2cm}
    \centering
    \includegraphics[width=6cm]{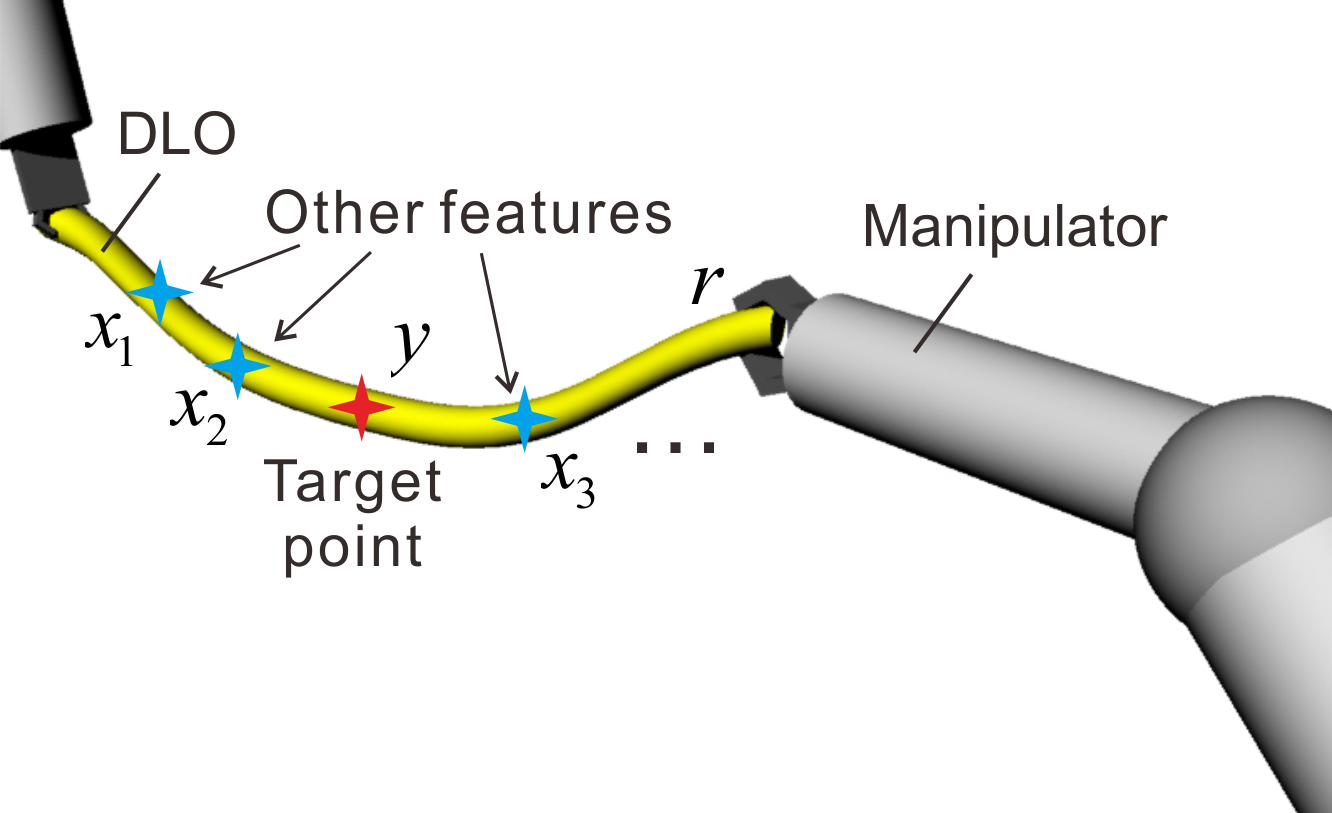}
    \caption{An illustration of robotic manipulation of DLOs. The robot grasps and manipulates the DLO to move the target point to the desired position. The overall shape of the DLO can be represented with multiple features along the DLO, which can be measured by sensors.}
    \label{fig:dualArm}
    \vspace{-0.3cm}
\end{figure}

Consider an illustration of robotic manipulation of DLOs shown in Fig. \ref{fig:dualArm}, where the robot grasps and manipulates the DLO to achieve the desired task, by controlling the motion of its end effector. The end tip and also the overall shape of the DLO can be measured with sensors. In this paper, the manipulation task is simplified as moving the target point on the DLO to the desired position.

Then, the velocity of the target point on the DLO can be related to the velocity of the robot end effector using the Jacobian matrix. 
Compared to \cite{zhu_dual-arm_2018,lagneau_automatic_2020,zhu2019_3dDeformable,navarro2016Automatic,navarro2018fourier}, the overall shape of the DLO is considered in the Jacobian matrix as 
\begin{equation} \label{Jacob1}
    \dot{\bm y}=\bm J(\bm{\phi})\dot{\bm r}
\end{equation}
where $\bm{\phi}$ represents the overall shape of the DLO. It can be specifically represented as $\bm{\phi} \stackrel{\triangle}{=} [\bm{x}_1^T, \cdots, \bm{x}_m^T]^T$, where $\bm{x}_i \in \Re^l$ is the position of the $i^{\rm th}$ feature along the DLO and $m$ is the number of features, $\bm y \in \Re^l$ is the position of the target point on the DLO, $\bm r \in \Re^n$ is the pose of the robot end effector, $\bm J(\bm{\phi}) \in \Re^{l\times n}$ is the Jacobian matrix of the DLO with the shape $\bm{\phi}$, which is bounded.

Note that any point along the DLO can be set as the target point, and different points correspond to different Jacobian matrices. Equation (\ref{Jacob1}) can be extended to  features as $\dot{\bm x_i}=\bm J^{\bm x_i}(\bm{\phi})\dot{\bm r}$ where $\bm J^{\bm x_i}(\bm{\phi})$ is the Jacobian matrix for the $i^{\rm th}$ feature. For the sake of illustration, the target point in this paper is referred to as $\bm y$ and the Jacobian matrix is referred to as $\bm{J}(\bm{\phi})$. Note that the target point in specific manipulation tasks can be defined as one of the features (i.e. $\bm y \stackrel{\triangle}{=} \bm x_i$, $\bm{J}(\bm{\phi}) \stackrel{\triangle}{=} \bm J^{\bm x_i}(\bm{\phi})$).

The Jacobian matrix $\bm{J}(\bm{\phi})$ is dependent on the deformation model of DLOs, which may involve many parameters \cite{ogden1997non,henrich2012robot,sadd2009elasticity}. Although the parameters can be experimentally calibrated, any modeling bias during the calibration compromises the accuracy of the model. Since the material and dimension vary significantly among different DLOs, and the length of the same DLO also changes when it is manipulated, the model-based calibration methods are not effective for accommodating variations in the material, dimension, and length of the DLO.

When the deformation model is unknown, the Jacobian matrix is also unknown. In this paper, the unknown Jacobian matrix will be estimated in both the offline (before manipulation) and online (during manipulation) manners. The estimated Jacobian matrix, denoted as $\hat{\bm J}(\bm{\phi})$, will be employed in the control law to relate the velocity of the robot end effector to the velocity of the target point.

\begin{figure}[!tb]
    \vspace{0.2cm}
    \centering
    \includegraphics[width=8.5cm]{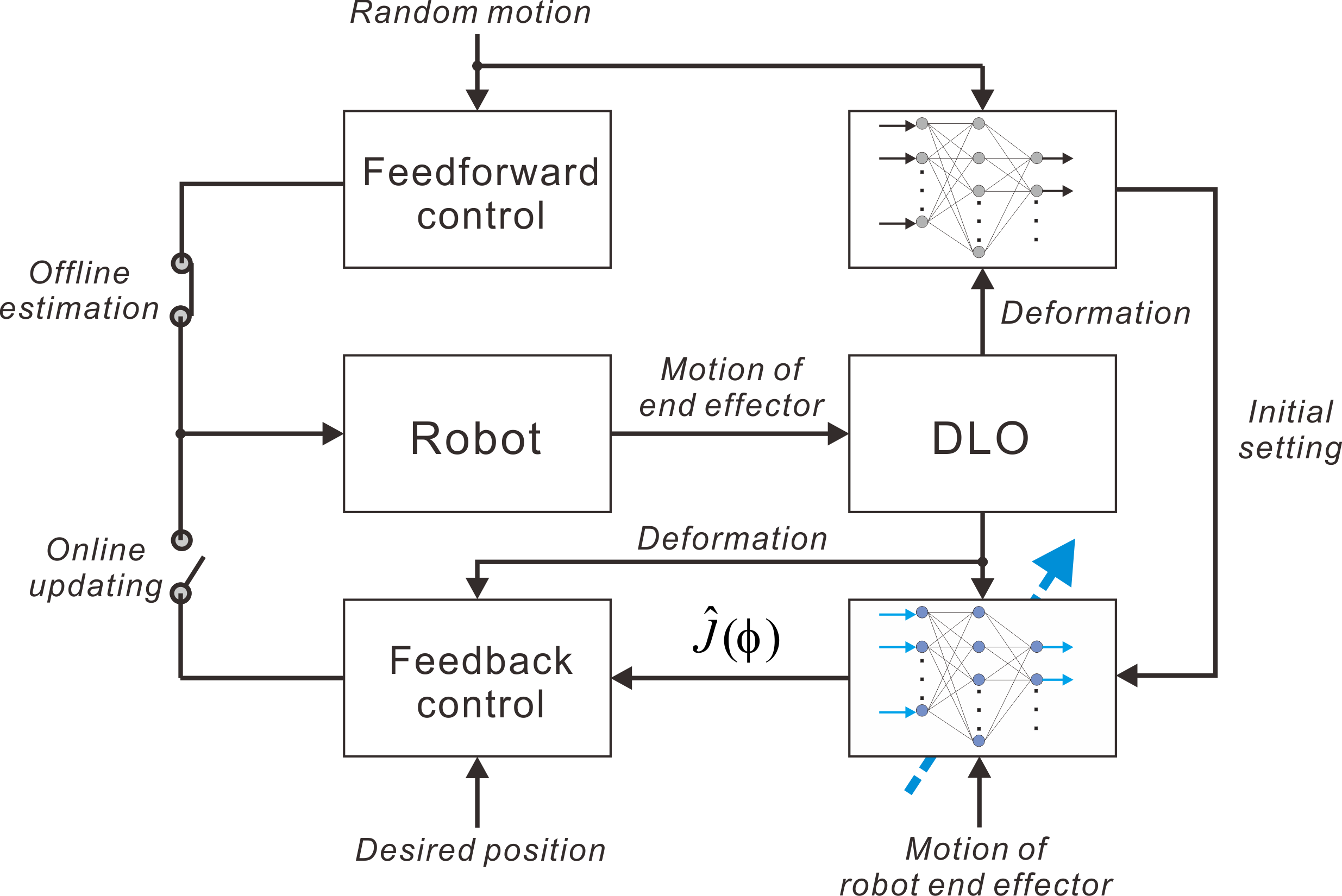}
    \caption{The proposed scheme consists of both the offline learning and the online learning, and the results obtained in the offline phase can be directly migrated to the online phase without any additional formatting. The system in this figure is in the offline phase.}
    \label{fig:whole_scheme}
    \vspace{-0.3cm}
\end{figure}

The control input is set as the velocity of the robot end effector as \cite{zhu_dual-arm_2018,lagneau_automatic_2020,zhu2019_3dDeformable,navarro2016Automatic,navarro2018fourier}
\begin{equation} \label{sysModel}
    \dot{\bm r}=\bm u 
\end{equation}
where $\bm{u}$ denotes the input. The structure of the proposed scheme is shown in Fig. \ref{fig:whole_scheme}. In the phase of offline learning, the dataset can be collected by controlling the robot end effector to follow a set of time-varying paths in an open-loop manner, while recording the variations of features. In the phase of online learning, the robot end effector is controlled to manipulate the target point to the desired position, by keeping updating the previously learnt model.

\section{Offline Modeling of DLOs}

Prior to the formal manipulation, a data-driven learning method is employed to obtain the initial model of DLOs. As the radial-basis-function neural network (RBFN) is commonly used in adaptive control and machine learning \cite{xli2014adpative}, the actual Jacobian matrix is represented with RBFN in this paper as
\begin{equation} \label{vecJ=W*theta}
    {\rm{vec}} (\bm{J}(\bm{\phi})) = \bm{W}\bm{\theta}(\bm{\phi})
\end{equation}
where $\bm{W}$ is the matrix of actual weights of the NN (which are unknown), $\bm{\theta}(\bm{\phi})$ represents the vector of activation functions. In addition, $\bm{\theta}(\bm{\phi})\hspace{-0.05cm}=\hspace{-0.05cm}[\theta_1(\bm{\phi}), \theta_2(\bm{\phi}), \cdots, \theta_q(\bm{\phi})]^T \in \Re^q $. An example of the activation function is the Gaussian radial function, that is
\begin{equation} \label{singleNeuron}
\theta_i(\bm{\phi})={\rm e}^{\frac{-||\bm{\phi} -\bm \mu_i||^2}{\sigma_i^2}}, \quad i =  1,\cdots , q
\end{equation}
where $\bm{\phi} = [\bm{x}_1^T, \cdots, \bm{x}_m^T]^T$ is the input vector of the NN. 

Equation (\ref{vecJ=W*theta}) can be decomposed as
\begin{equation}
    \bm{J}_i(\bm{\phi}) = \bm{W}_i\bm{\theta}(\bm{\phi})
\end{equation}
where $\bm{J}_i, (i=1 , \cdots , n)$ is the $i^{\rm th}$ column of the Jacobian matrix, and $\bm{W}_i$ is the ${((i-1)\times l+1)}^{\rm th}$ to ${(i\times l)}^{\rm th}$ rows of $\bm{W}$. Then (\ref{Jacob1}) can be written as
\begin{equation}
    \dot{\bm{y}} = \bm{J}(\bm{\phi})\dot{\bm{r}} = \sum_{i=1}^{n}\bm{J}_i(\bm{\phi})\dot{r}_i = \sum_{i=1}^{n}\bm{W}_i\bm{\theta}(\bm{\phi})\dot{r}_i
\end{equation}

The estimated Jacobian matrix is represented as 
\begin{equation} \label{estimatedJ}
    {\rm{vec}} (\hat{\bm{J}}(\bm{\phi})) = \hat{\bm{W}}\bm{\theta}(\bm{\phi})
\end{equation}
where $\hat{\bm{W}}$ is the matrix of estimated weights. The approximation error $\bm{e}_w$ is specified as 
\begin{equation} \label{ew}
\begin{aligned}
    \bm{e}_w &= \dot{\bm{y}} - \hat{\bm{J}}(\bm{\phi})\dot{\bm{r}}
    = (\bm{J}(\bm{\phi})  -\hat{\bm{J}}(\bm{\phi}))\dot{\bm{r}} \\
    &= \sum_{i=1}^{n}\bm{W}_i\bm{\theta}(\bm{\phi})\dot{r}_i - \sum_{i=1}^{n}\hat{\bm{W}}_i\bm{\theta}(\bm{\phi})\dot{r}_i = \sum_{i=1}^{n}\Delta{\bm{W}}_i\bm{\theta}(\bm{\phi})\dot{r}_i
\end{aligned}
\end{equation}

\begin{figure}[!tb]
    \vspace{0.2cm}
    \centering
    \includegraphics[width=6.3cm]{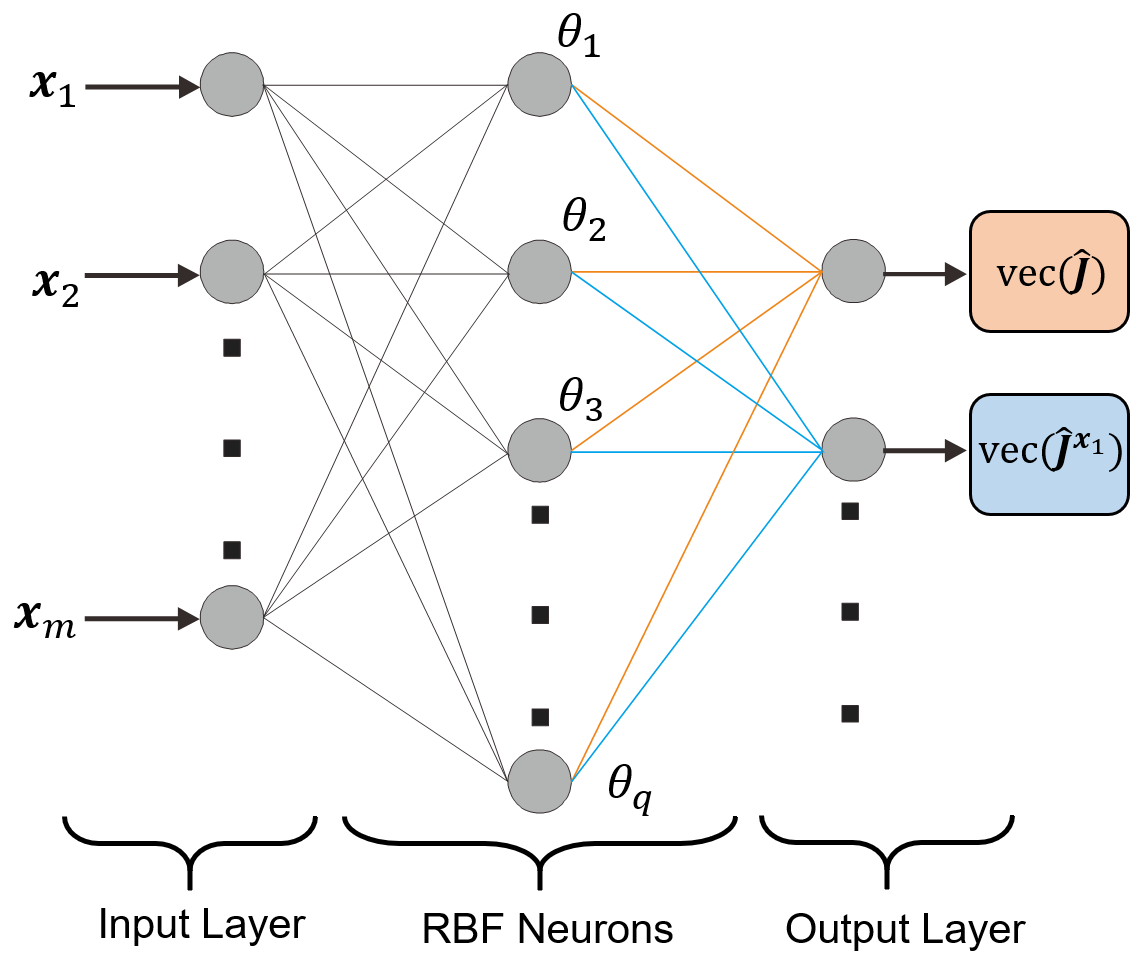}
    \caption{The structure of the RBFN for modeling DLOs. The NNs take the overall shape of the DLO (the positions of the features along the DLO) as the input and finally output the estimated Jacobian matrices relating the velocities of points on the DLO to the velocity of robot end effector. The NNs for the Jacobian matrices of the target point and all features can be trained concurrently with the collected data.}
    \label{fig:NNstructure}
    \vspace{-0.3cm}
\end{figure}

In the offline learning phase, the robot end effector is controlled to move in an open-loop manner to collect the training dataset first, which contains $\bm x_i, \dot{\bm{x}_i}, \dot{\bm{r}}, \dot{\bm{y}}, (i=1,\cdots,m)$. As these data can be obtained concurrently, NNs are trained for all the Jacobian matrices of $\bm{J}(\bm{\phi})$ and $\bm{J}^{\bm x_i}(\bm{\phi}), (i=1,\cdots,m)$ for the sake of efficiency. After all the Jacobian matrices are well estimated by the end of the offline phase, the target point can also be chosen as one of the other features, by replacing the Jacobian matrix with the corresponding one. The structure of the whole NNs is shown in Fig. \ref{fig:NNstructure}. 

Next, the RBF network is trained with the collected data. Considering the noise and the outliers in the data, the smooth $L1$ loss is used for training, which is specified as 
\begin{equation} \label{offline_loss}
L(\hat{\bm{J}}(\bm{\phi})) = \sum_{j=1}^{l} {L_j}
\end{equation}
where
\begin{equation} \label{offline_loss_2}
L_j = \left\{ \begin{array}{rl}
0.5(e_{wj})^2 / \beta  & , |e_{wj}| < \beta \\ 
|e_{wj}|-0.5\beta  & , \rm{otherwise}
\end{array}\right.
\end{equation}
where $e_{wj}$ is the $j^{\rm th}$ element of $\bm e_w$.

The k-means clustering on sampled training data is used to calculate the initial value of $\bm{\mu}_i$ and $\sigma_i, (i = 1, \cdots,q)$. Then, all parameters including $\bm{\mu}_i$, $\sigma_i$ and $\hat{\bm{W}}$ are updated by the back propagation of the loss in (\ref{offline_loss}). The {\em Adam} optimizer \cite{kingma2014adam} is used for training. 
Note that the estimated parameters of the NN in the offline phase can be directly migrated to the online phase. In the online phase, the parameters will be further updated to meet the specific manipulation task.

\section{Adaptive Control with Online Learning}
Due to insufficient training data or changes of the parameters of DLOs, the approximation errors may still exist by the end of the offline learning. In this section, an adaptive control scheme is proposed for robotic manipulation of DLOs, by treating the estimated Jacobian matrix as an initial approximation then further updating it during manipulation.

The control input is specified as
\begin{equation} \label{control_law}
\bm u=\hat{\bm J}^{\dagger}(\bm{\phi})(\dot{\bm y}_{d}-\bm K_p\Delta\bm y)
\end{equation}
where $\hat{\bm J}^{\dagger}(\bm{\phi})$ is the Moore-Penrose pseudo-inverse of the estimated Jacobian matrix, and it is assumed that $\hat{\bm J}(\bm{\phi})$ is full row rank and $\hat{\bm J}^{\dagger}(\bm{\phi})$ always exists. In addition, $\Delta \bm{y} = \bm{y} - \bm{y}_d$ where $\bm y_d \in \Re^l$ specifies the desired position of the target point, and $\bm{K}_p \in \Re^{l \times l}$ is the control gain, which is diagonal and positive definite. 

The online updating law of the $j^{\rm th}, (j=1 ,\cdots, l)$ row of $\hat{\bm W_i}$ is specified as 
\begin{equation} \label{control_update}
\dot{\hat{\bm w}}_{ij}^T = \dot{r}_i \bm L_i\bm\theta(\bm{\phi}) (\Delta y_{j} + \lambda e_{wj})
\end{equation}
where $\Delta y_j$ is the $j^{\rm th}$ element of the vector $\Delta \bm{y}$, and $\bm L_i \in \Re^{q\times q}$ is a positive-definite matrix, and $\lambda$ is a positive scalar. 

The proposed control scheme by (\ref{control_law}) and (\ref{control_update}) has several advantages as
\begin{enumerate}
    \item[-] The well estimated weights of the NN in the offline phase can be directly migrated as the initial values in the online phase. 
    
    \item[-] It allows the robot to manipulate the DLO by following the desired path (i.e. $\bm y_d$) and also update the unknown deformation model concurrently.
    
    \item[-] The update is driven by both the approximation errors (i.e. $\bm e_w$) and the task errors (i.e. $\Delta\bm y$), ensuring the faster convergence of the weights of the NN. 
\end{enumerate}

Substituting (\ref{control_law}) into (\ref{sysModel}), the closed-loop equation is obtained as
\begin{equation} \label{control_loop_equation_1}
\dot{\bm r}=\hat{\bm J}^{\dagger}(\bm{\phi})(\dot{\bm y}_{d}-\bm K_p\Delta\bm y)
\end{equation}
Multiplying both sides of (\ref{control_loop_equation_1}) with $\hat{\bm J}(\bm{\phi})$, we have
\begin{equation} \label{control_loop_equation_2}
\hat{\bm J}(\bm{\phi})\dot{\bm r}=\dot{\bm y}_{d}-\bm K_p\Delta\bm y
\end{equation}
Note that
\begin{equation} \label{control_loop_equation_3}
\hat{\bm J}(\bm{\phi})\dot{\bm r}=\hat{\bm J}(\bm{\phi})\dot{\bm r}-{\bm J}(\bm{\phi})\dot{\bm r}+{\bm J}(\bm{\phi})\dot{\bm r} = -\sum_{i=1}^{n} \Delta\bm{W}_i\bm{\theta}(\bm{\phi})\dot{r}_i + \dot{\bm y}
\end{equation}
Substituting (\ref{control_loop_equation_3}) into (\ref{control_loop_equation_2})  and using (\ref{ew}) yields
\begin{equation} \label{control_loop_equation_4}
\bm e_w = \Delta\dot{\bm y}+\bm K_p\Delta\bm y
\end{equation}
That is, the approximation errors are now expressed in terms of the task errors. Hence the convergence of $\bm e_w$ to zero naturally guarantees the realization of manipulation task.

To prove the convergence, the Lyapunov-like candidate is given as
\begin{equation} \label{V2}
V=\frac{1}{2}\Delta\bm y^T\Delta\bm y + \frac{1}{2}\sum\limits_{i=1}^n\sum\limits_{j=1}^l\hspace{0.1cm}\Delta\bm w_{ij}\bm L_i^{-1}\Delta\bm w_{ij}^T
\end{equation}
Differentiating (\ref{V2}) with respect to time and substituting (\ref{control_loop_equation_4}) into it, we have
\begin{equation} \label{dotV2}
\begin{aligned}
\dot V&= \Delta\bm y^T\Delta\dot{\bm y}-\sum\limits_{i=1}^n\sum\limits_{j=1}^l\Delta\bm w_{ij}\bm L_i^{-1}\dot{\hat{\bm w}}_{ij}^T\\
&= \Delta\bm y^T(\bm e_w -\bm K_p\Delta\bm y)
-\sum_{i=1}^n\sum\limits_{j=1}^l\Delta\bm w_{ij}\bm L_i^{-1}\dot{\hat{\bm w}}_{ij}^T\\
&=-\Delta\bm y^T\bm K_p\Delta\bm y+\Delta\bm y^T\bm e_w -\sum\limits_{i=1}^n\sum\limits_{j=1}^l\Delta\bm w_{ij}\bm L_i^{-1}\dot{\hat{\bm w}}_{ij}^T
\end{aligned}
\end{equation}
Next substituting the update law (\ref{control_update}) into (\ref{dotV2}) and using (\ref{ew}), we have
\begin{equation} \label{dotV2_2}
\begin{aligned}
    \dot V &=  -\Delta\bm y^T\bm K_p\Delta\bm y+\Delta\bm y^T\bm e_w \\
    & \quad -\sum\limits_{i=1}^n\sum\limits_{j=1}^l\Delta\bm w_{ij}\bm L_i^{-1} [\dot{r}_i \bm L_i\bm\theta(\bm{\phi}) (\Delta y_{j} + \lambda e_{wj})] \\
    &= -\Delta\bm y^T\bm K_p\Delta\bm y + \Delta\bm y^T\bm e_w - \bm e_w^T \Delta\bm y - \lambda \bm{e}_w^T \bm{e}_w \\
    &= -\Delta\bm y^T\bm K_p\Delta\bm y - \lambda \bm{e}_w^T \bm{e}_w \leq 0
\end{aligned}
\end{equation}
Since $V>0$ and $\dot V\leq 0$, $V$ is bounded. The boundedness of $V$ ensures the boundedness of $\Delta\bm w_{ij}$ and $\Delta\bm y$. From (\ref{control_loop_equation_1}), $\dot{\bm r}$ is also bounded. The boundedness of $\dot{\bm r}$ ensures the boundedness of $\dot{\bm y}$ from (\ref{Jacob1}). Hence,  $\Delta\bm y$ is uniformly continuous. 
From (\ref{dotV2_2}), it can be shown that $\Delta\bm y \in L_2(0, +\infty)$. Then, it follows \cite{arimoto1996control} that $\Delta\bm y \rightarrow \bm 0$ as $t\rightarrow\infty$. Therefore, 
the manipulation task is achieved.

\section{Simulation Studies}

Simulations are carried out to study the performance of the proposed method. The simulation environment is built in Unity \cite{unity}, a 3D game development platform. The simulation of the DLO is based on Obi \cite{obi}, a unified particle physics for Unity in which realistic deformable objects can be created, such as ropes, cloth, fluids and other softbodies. The ROS \cite{quigley2009ros} and ROS\# \cite{RosSharp} are used for the communication between the physical simulation in Unity and the control program written in Python scripts. All the simulation and computation are done on a Ubuntu 18.04 desktop (CPU: Intel i7-10700, GPU: Nvidia GeForce RTX 3070, RAM: 16GB).

The simulation scene is shown in Fig. \ref{fig:simulation_scene}. The DLO is modeled with the rod blueprint in Obi Rope package, which is built by chaining oriented particles using stretch/shear and bend/twist constraints. Its length is about 0.5m and radius is about 5mm. The blue points represent the 10 features along the DLO, and the red point represents the target point. Note that one of the features is chosen as the target point for convenience, so the red point is also a feature. The virtual green point in the simulation scene represents the desired position of the target point. The left end of the DLO is grasped and fixed by one robot, and the right end of the DLO is grasped by another robot and the linear velocity of the robot end effector is treated as the control input. This paper considers the positions of features and target point in 3-D world coordinate system, where the parameters are set as $l=3, n=3, m=10$ in following simulations.

\subsection{Offline Learning}

In the offline phase of modeling, the unknown Jacobian matrix was approximated with the proposed NN. First, the training data was collected by controlling the robot end effector to continuously move in the workspace. In each time period $\Delta T$, a desired position was randomly set in the workspace at the beginning. The end effector was controlled to 
reach the desired position at the end of $\Delta T$. The data of $\bm x_i, \dot{\bm{x}_i}, \dot{\bm{r}}, (i=1,\cdots,m)$ in the process were recorded for the subsequent training. The velocities were obtained by differentiating the corresponding positions. The NN trained with more data would have better modeling accuracy, but more time would be required to collect the data.

A RBFN with 256 neurons in the middle layer (i.e. $q=256$) was trained to model the Jacobian matrices of all the features along the DLO. The PyTorch \cite{pytorch} with CUDA support was used for the implementation of the offline training. The training data was adjusted to an appropriate range which would benefit the training of NN, and 
the $\beta$ in (\ref{offline_loss_2}) was set as $1.0$.

\begin{figure*} [!tbh]
  \vspace{0.2cm}
  \centering 
  \subfigure[]{ 
    \includegraphics[width=16cm]{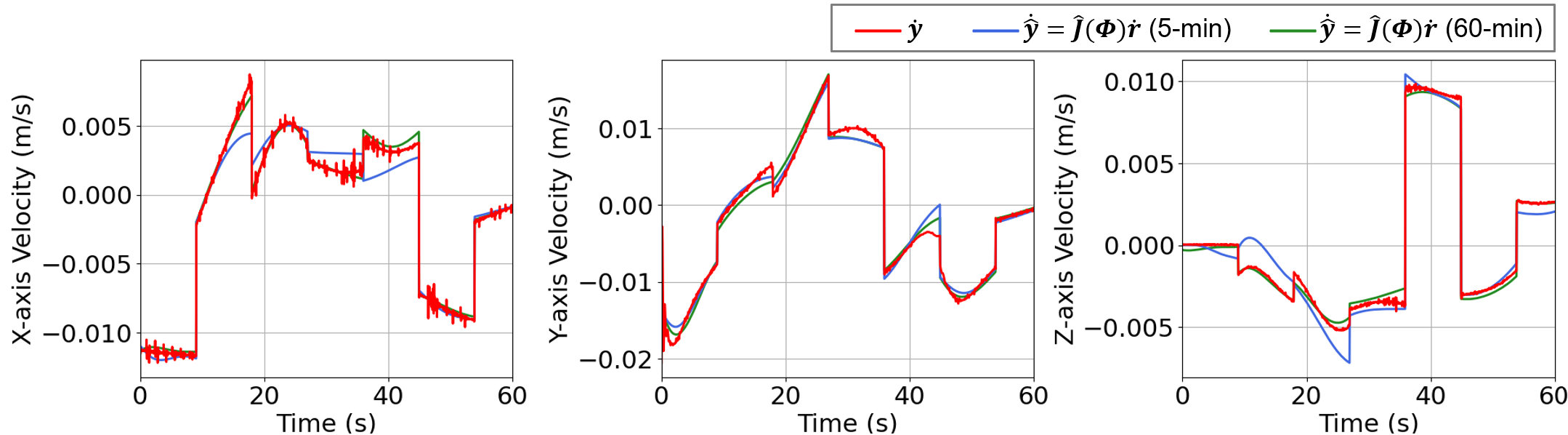} 
  } 
  \hspace{-2cm}
  \subfigure[]{ 
    \includegraphics[width=16cm]{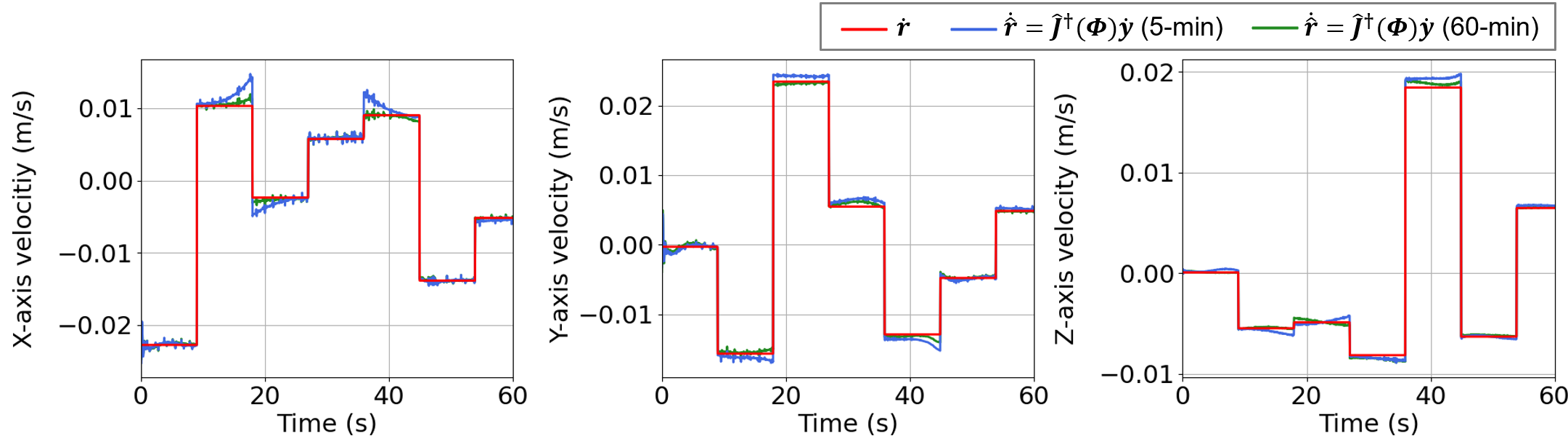} 
  } 
  \vspace{-0.3cm}
  \caption{ The modeling performance of the offline-trained NNs on 1-minute test data. The test data were collected in the same way as the training data. (a) Measured / predicted velocities of the target point:  The red line represents the measured velocities of the target point (i.e. $\dot{\bm y}$). The blue / green line represents the predicted velocities of it using the estimated Jacobian matrix output by the NN trained with 5-minute / 60-minute data (i.e. $ \dot{\hat{\bm y}} = \hat{\bm J}(\bm{\phi}) \dot{\bm r}$).
  (b) Measured / predicted velocities of the robot end effector: The red line represents the velocities of the robot end effector (i.e. $\dot{\bm r}$). The blue / green line represents the predicted velocities of it using the inverse of the estimated Jacobian matrix output by the NN trained with 5-minute / 60-minute data (i.e. $ \dot{\hat{\bm r}} = \hat{\bm J}^{\dagger}(\bm{\phi}) \dot{\bm y}$).
  } 
  \label{fig:offline_result}
  \vspace{-0.3cm}
\end{figure*}

To test how the amount of training data would influence the performance of the NN, two NNs were trained with 5-minute data and 60-minute data separately. Fig. \ref{fig:offline_result} shows the performance of the two trained NNs on the testset of another 1-minute data. In this figure, the target point was set as the fifth feature on the DLO. First, the comparison between the measured velocities of the target point (i.e. $\dot{\bm y}$) and the predicted velocities of it using the estimated Jacobian matrix (i.e. $ \dot{\hat{\bm y}} = \hat{\bm J}(\bm\phi) \dot{\bm r}$) is shown. Then, the comparison between the velocities of the robot end effector (i.e. $\dot{\bm r}$) and the predicted velocities of it using the inverse of the estimated Jacobian matrix (i.e. $ \dot{\hat{\bm r}} = \hat{\bm J}^{\dagger}(\bm\phi) \dot{\bm y}$) is also shown. Notice that limited by the accuracy of the DLO simulator, the measured velocities of the target point $\dot{\bm y}$ obtained by differentiating its positions contained noise, which also made the calculated $\dot{\hat{\bm r}}$ look unsmooth. It is illustrated that the estimated Jacobian matrix output by the NN trained with only 5-minute data can describe the relationship between the velocity of the target point and the velocity of robot end effector with sufficient accuracy, 
and the 60-minute training data can enable the NN to output a more accurate Jacobian matrix.

\subsection{Manipulation with Online Learning}

\begin{figure} [!tb]
  \centering 
  \subfigure[]{ 
    \includegraphics[width=4.0cm]{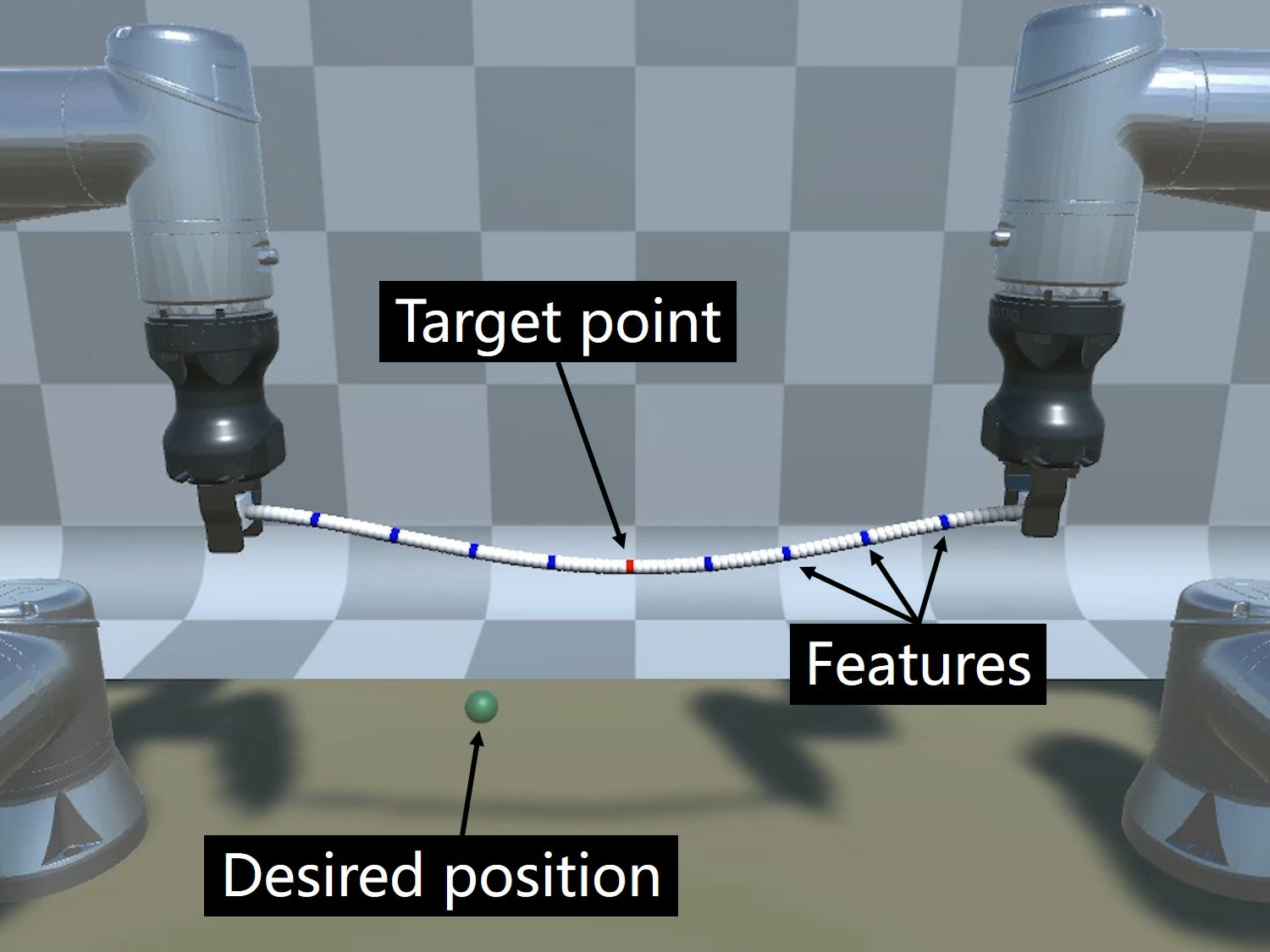} 
    \label{fig:simulation_scene}
  } 
  \hspace{-0.2cm}
  \subfigure[]{ 
    \includegraphics[width=4.0cm]{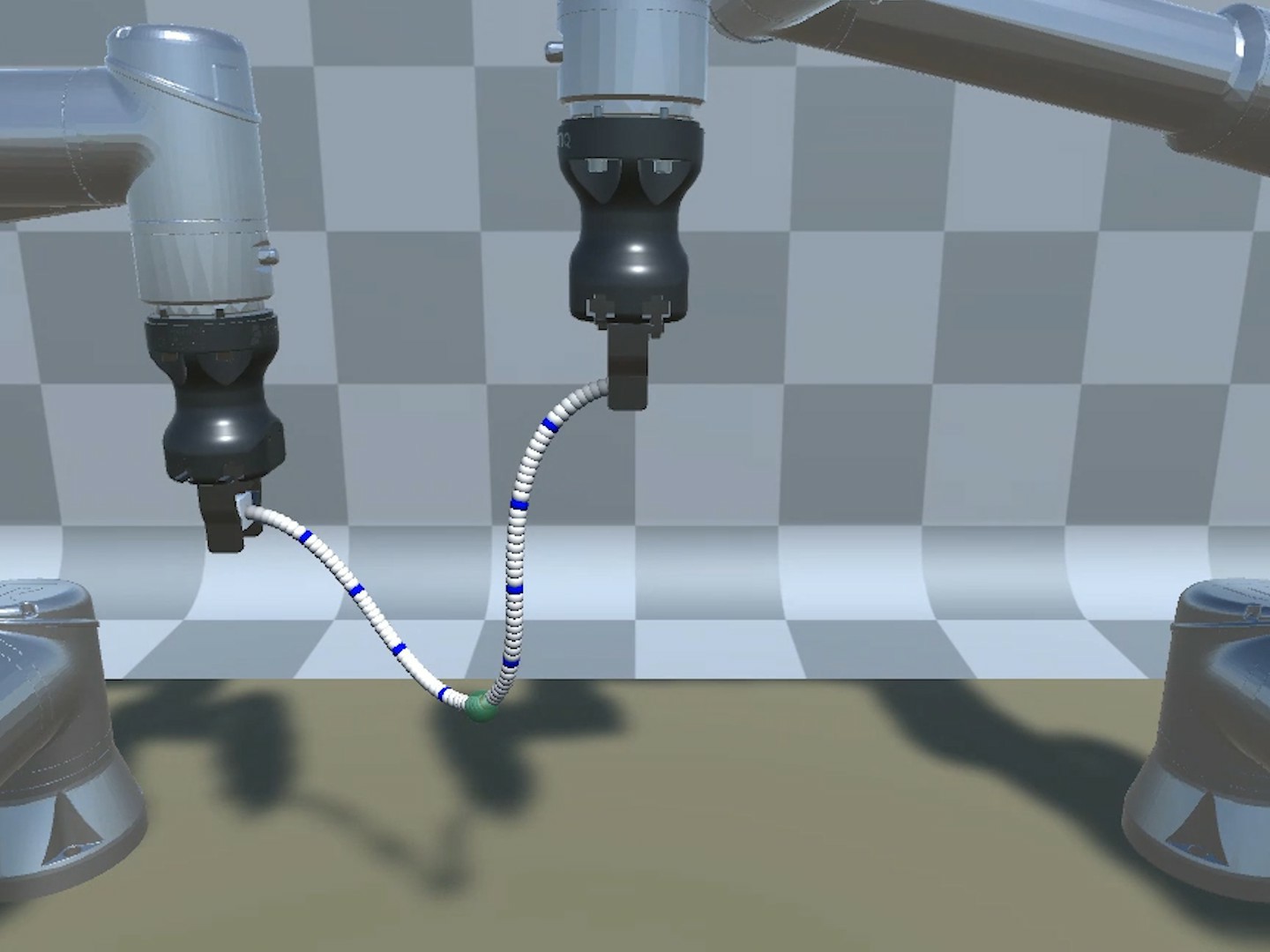} 
  }
  \caption{Snapshots of the manipulation task 1. The robot end effector was controlled to move the target point on the DLO to the desired position. (a) The initial state. (b) The target point reached the desired position.} 
  \label{fig:exp1}
  \vspace{-0.3cm}
\end{figure}

\begin{figure}[!tb]
    \centering
    \includegraphics[width=6cm]{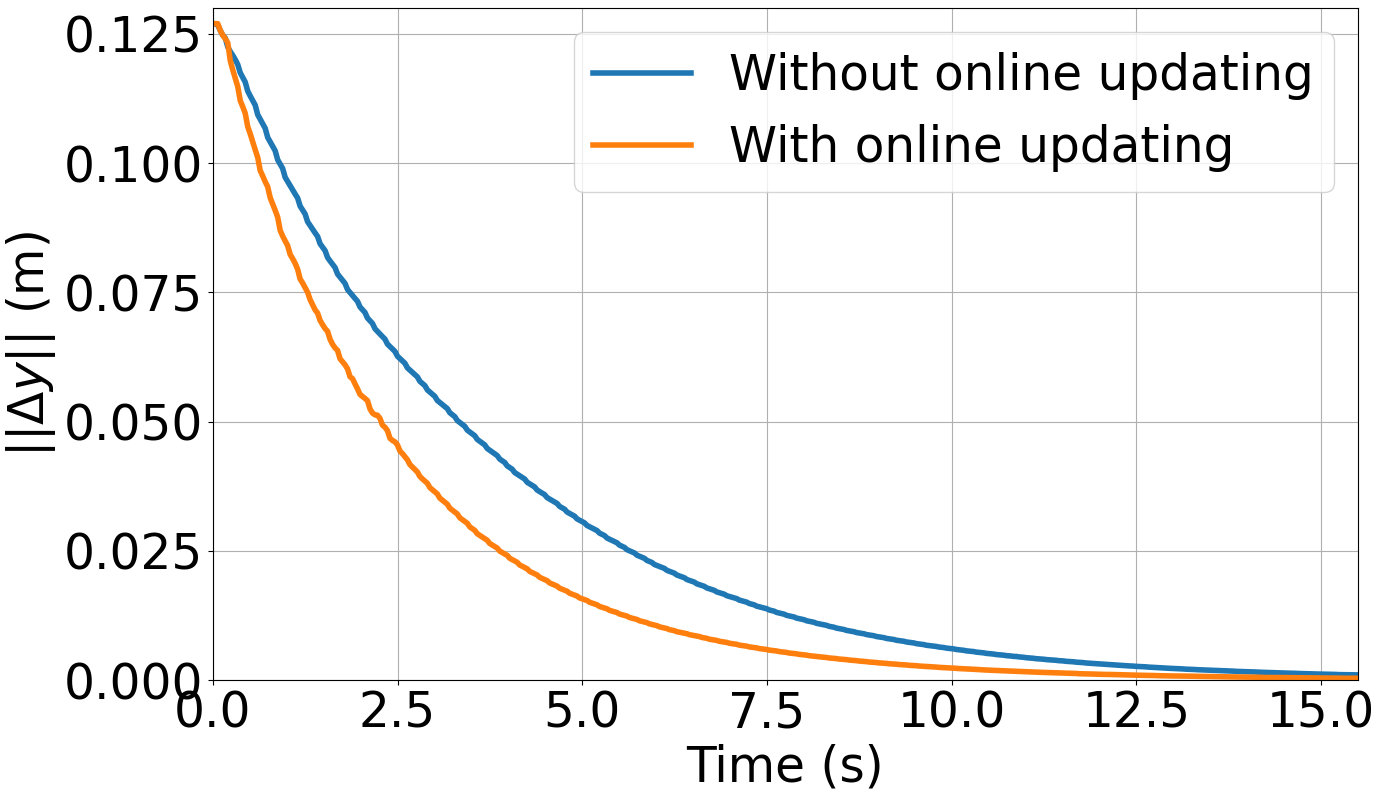}
    \caption{The comparison of the two manipulation processes of task 1, with or without the online updating, and $\| \Delta \bm y \|$ is the distance between the actual position and the desired position of the target point.}
    \label{fig:online_origin}
    \vspace{-0.3cm}
\end{figure}

\begin{figure*} [!tb]
  \centering 
  \subfigure[]{ 
    \includegraphics[width=3.4cm]{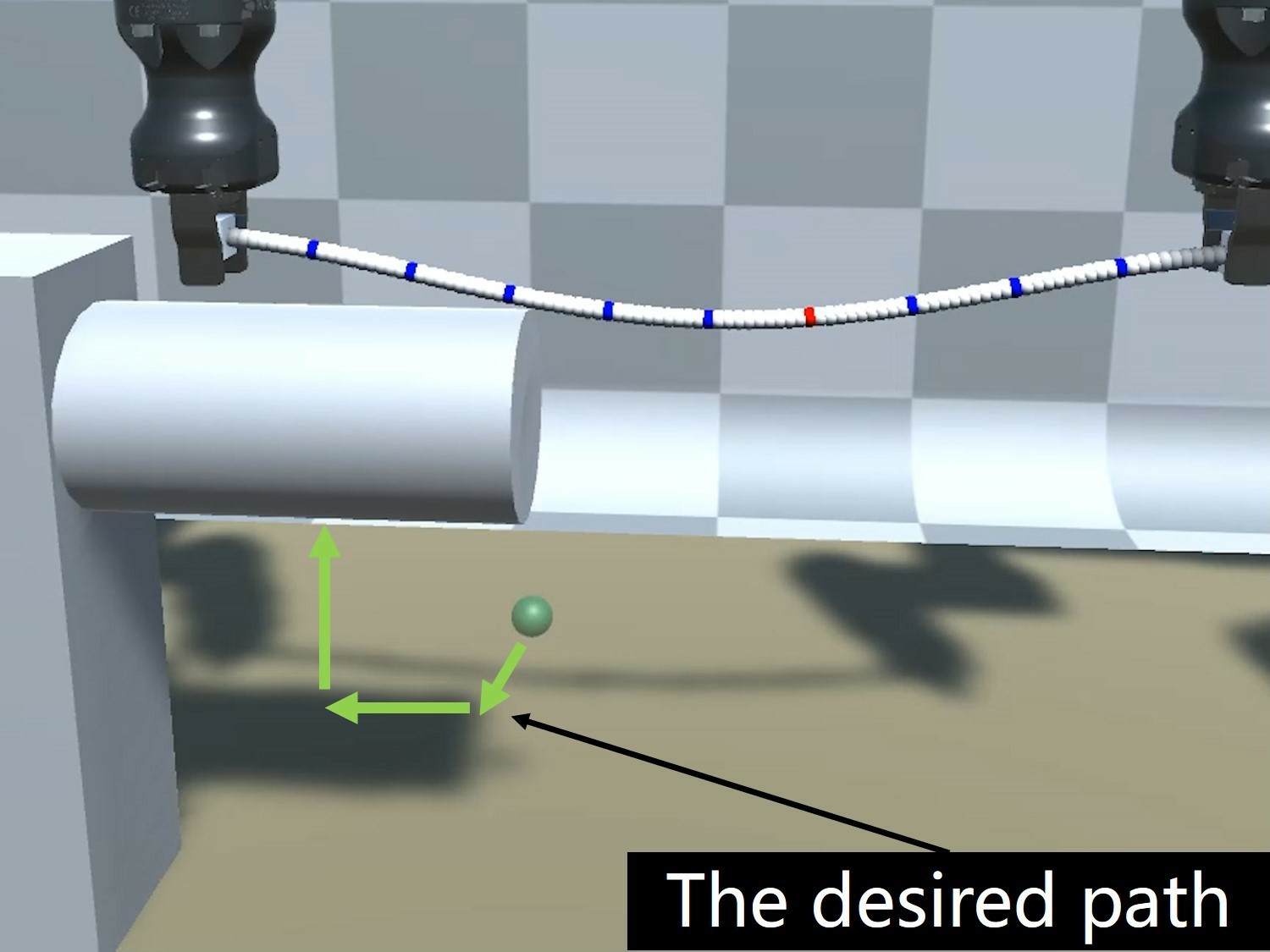} 
  }
  \hspace{-0.4cm}
  \subfigure[]{ 
    \includegraphics[width=3.4cm]{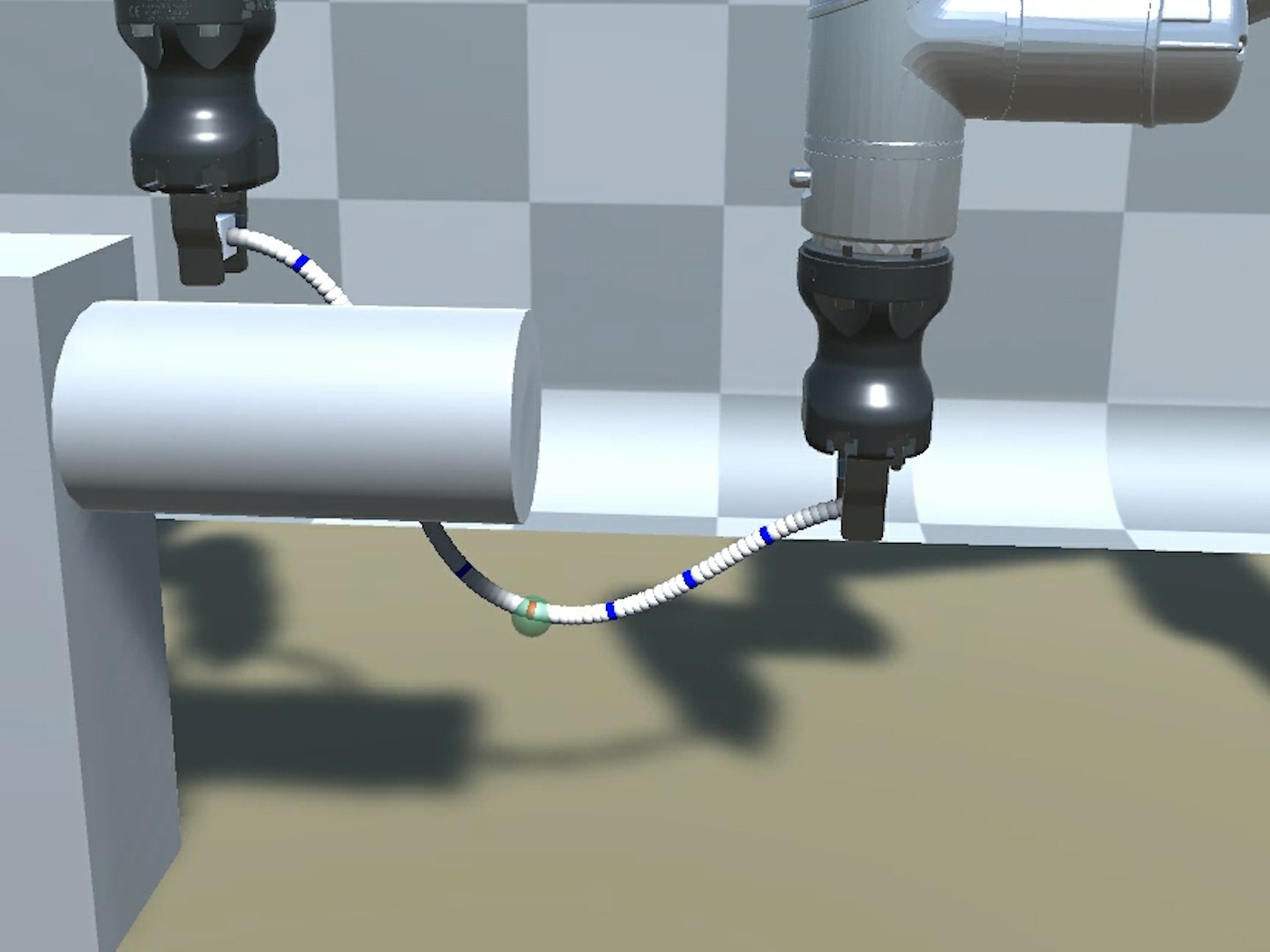} 
  } 
  \hspace{-0.4cm}
  \subfigure[]{ 
    \includegraphics[width=3.4cm]{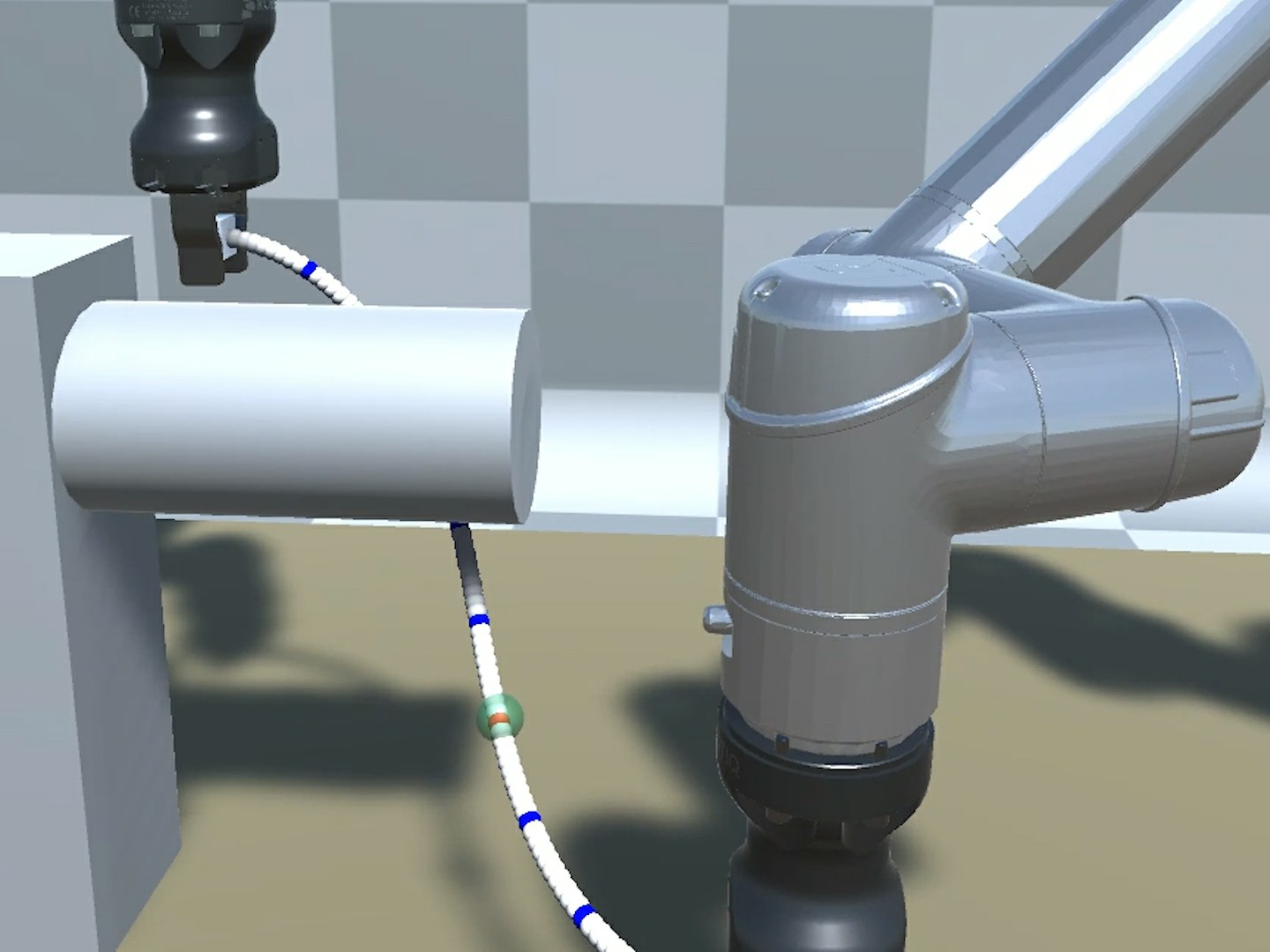} 
  }
  \hspace{-0.4cm}
  \subfigure[]{ 
    \includegraphics[width=3.4cm]{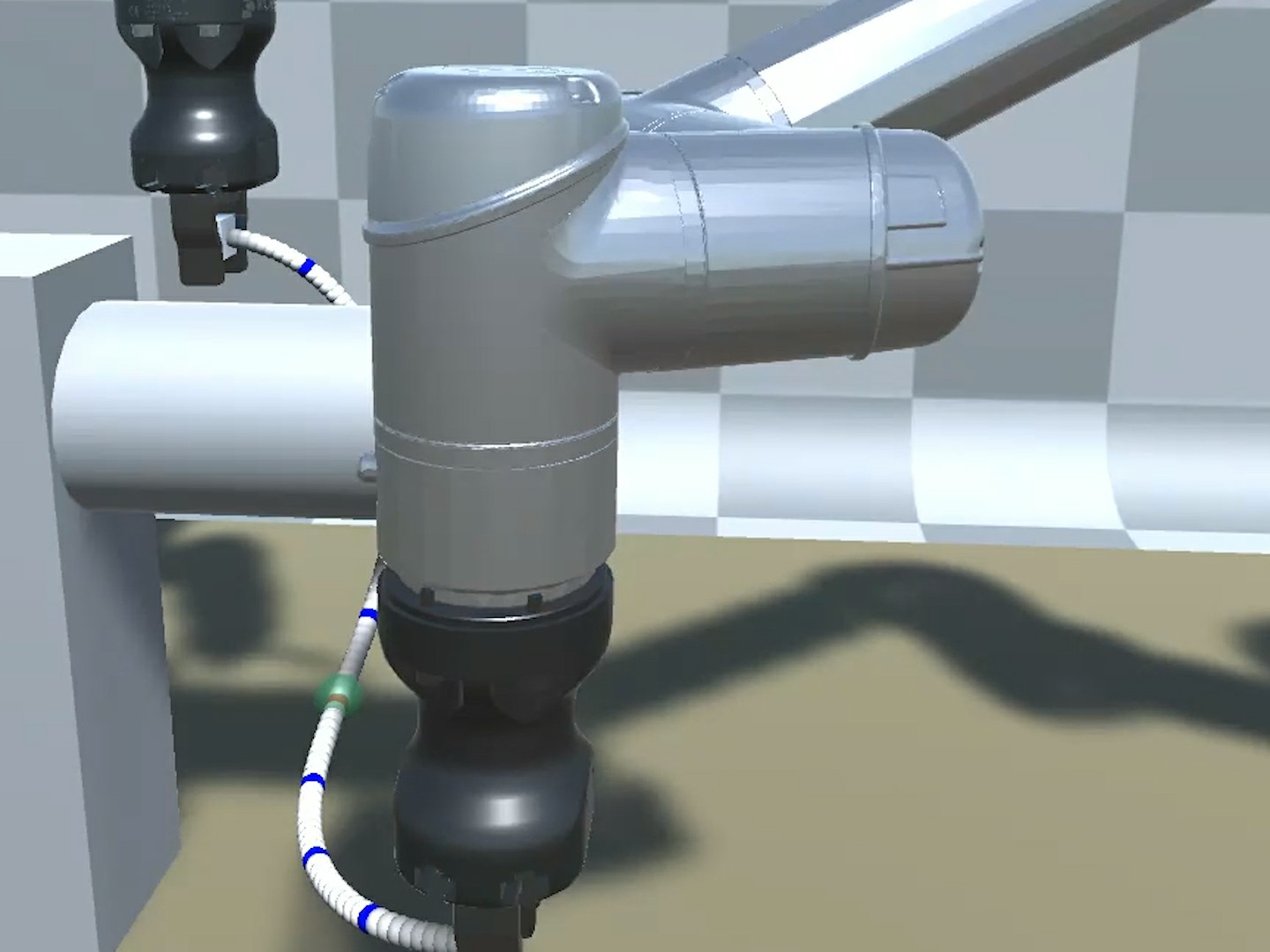} 
  } 
  \hspace{-0.4cm}
  \subfigure[]{ 
    \includegraphics[width=3.4cm]{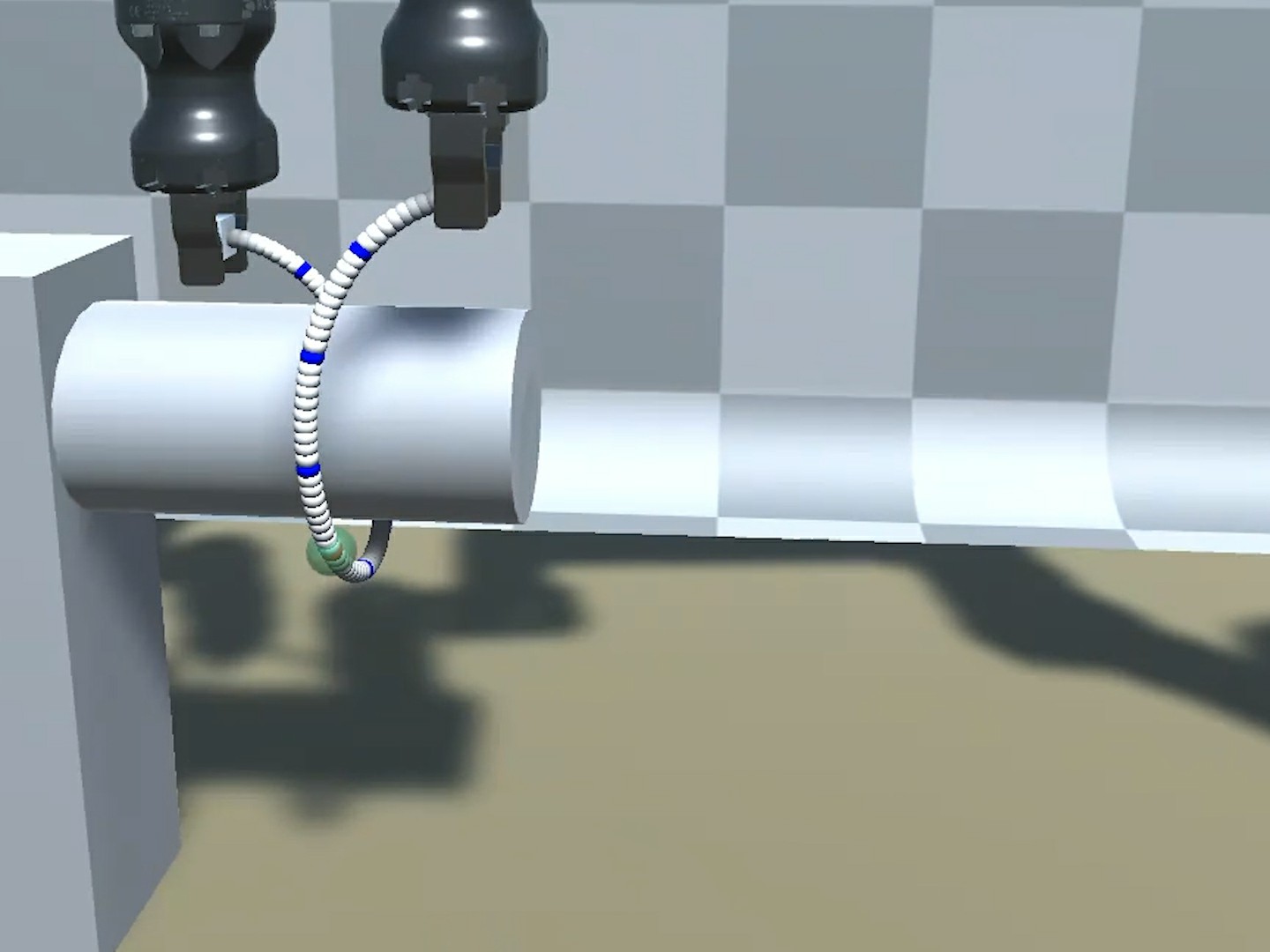} 
  } 
  \caption{Snapshots of the manipulation task 2. The DLO was manipulated to encircle the cylinder. A desired path was manually defined. The target point was controlled to follow the desired path. (a) $t = 0 \hspace{0.2mm} {\rm s}$: The initial state and the whole desired path. 
  (b) $t = 10  \hspace{0.2mm} {\rm s}$: Reached the first desired position.
  (c) $t = 15  \hspace{0.2mm} {\rm s}$: Followed the desired path.
  (d) $t = 21  \hspace{0.2mm} {\rm s}$: Followed the desired path.
  (e) $t = 28  \hspace{0.2mm} {\rm s}$: Reached the final desired position. 
  The task was completed. }
  \label{fig:exp2cylinder}
\end{figure*}

\begin{figure*} [!tb]
  \centering 
  \subfigure[]{ 
    \includegraphics[width=3.4cm]{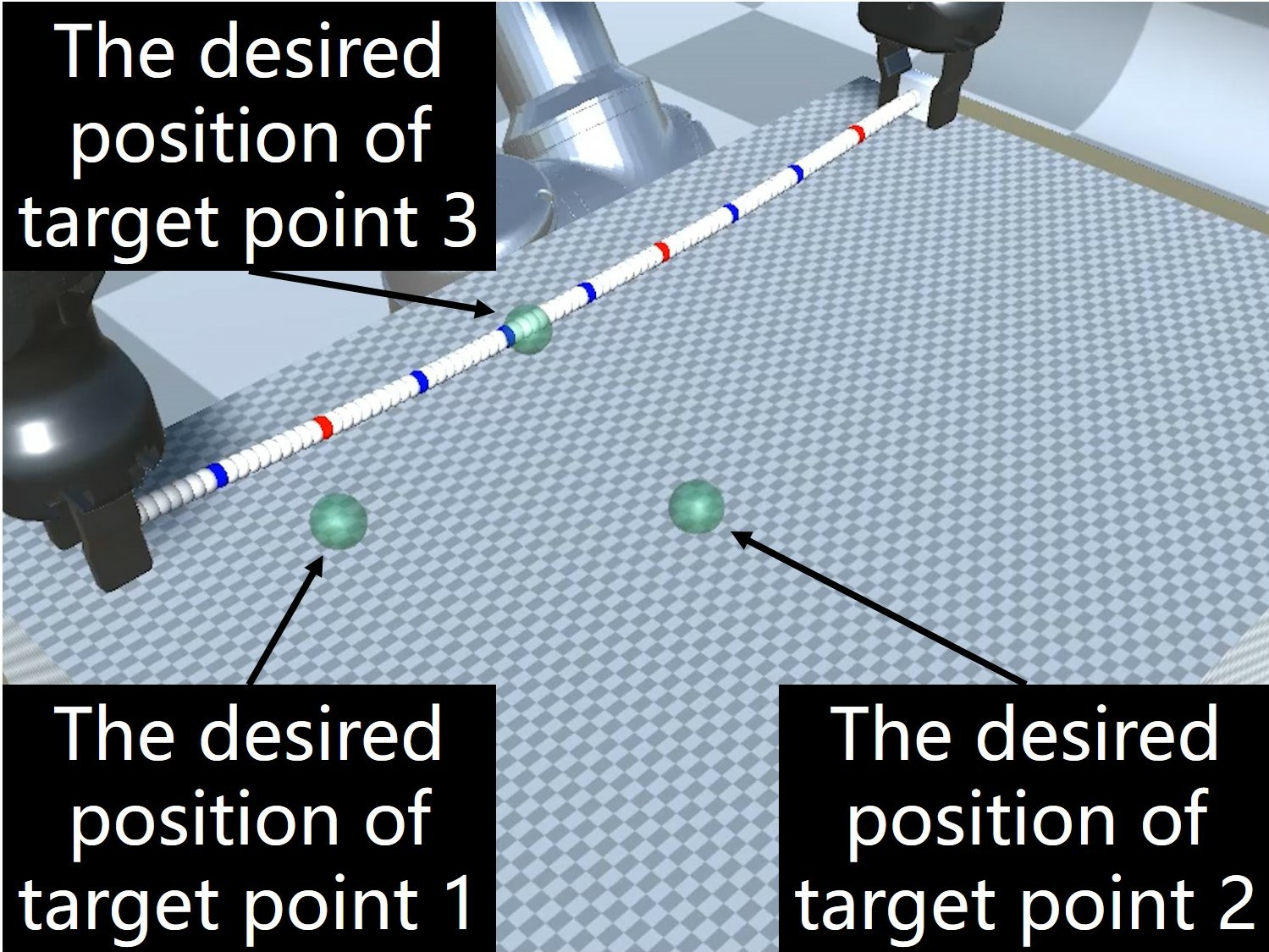} 
  }
  \hspace{-0.4cm}
  \subfigure[]{ 
    \includegraphics[width=3.4cm]{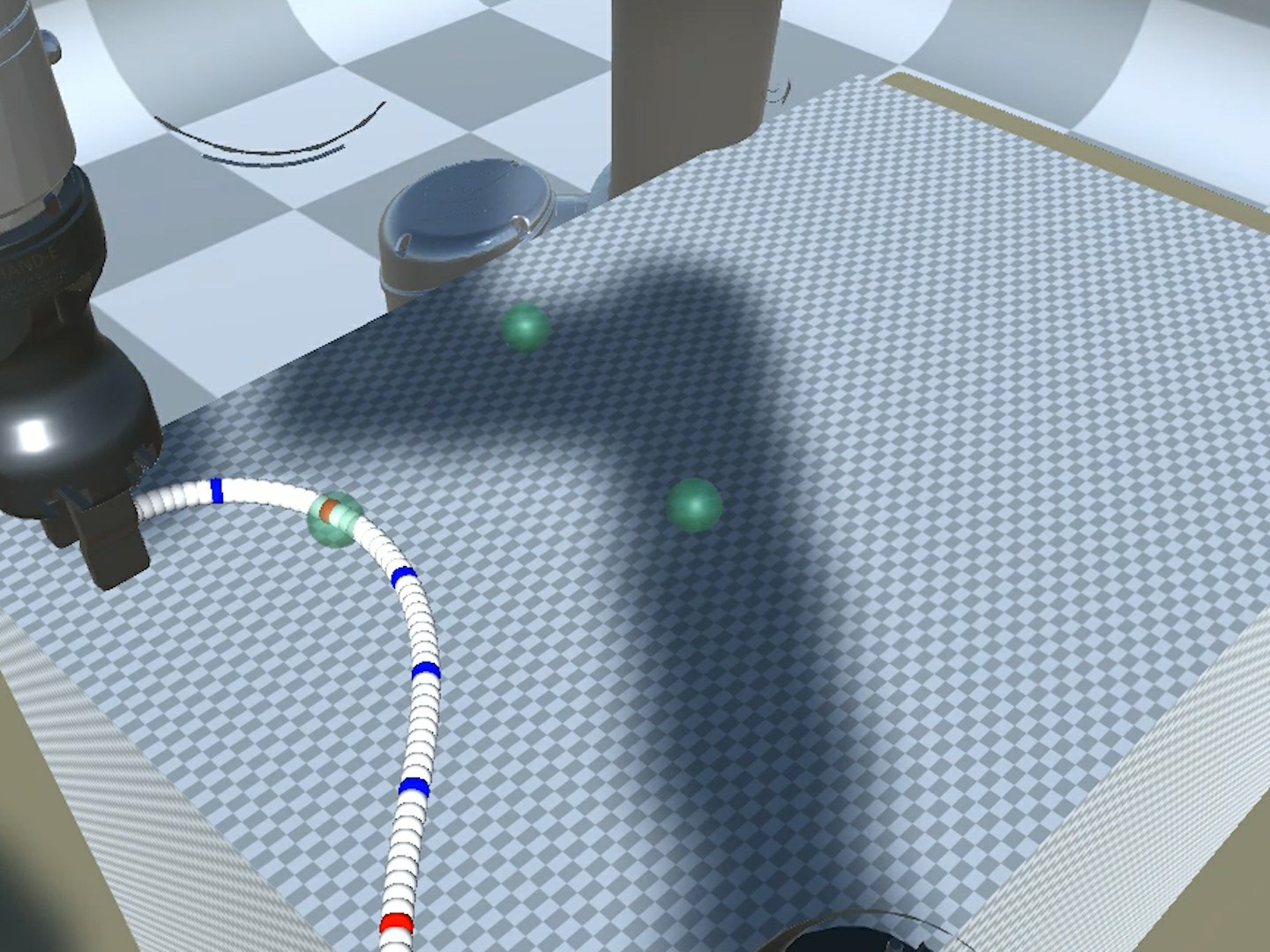} 
  } 
  \hspace{-0.4cm}
  \subfigure[]{ 
    \includegraphics[width=3.4cm]{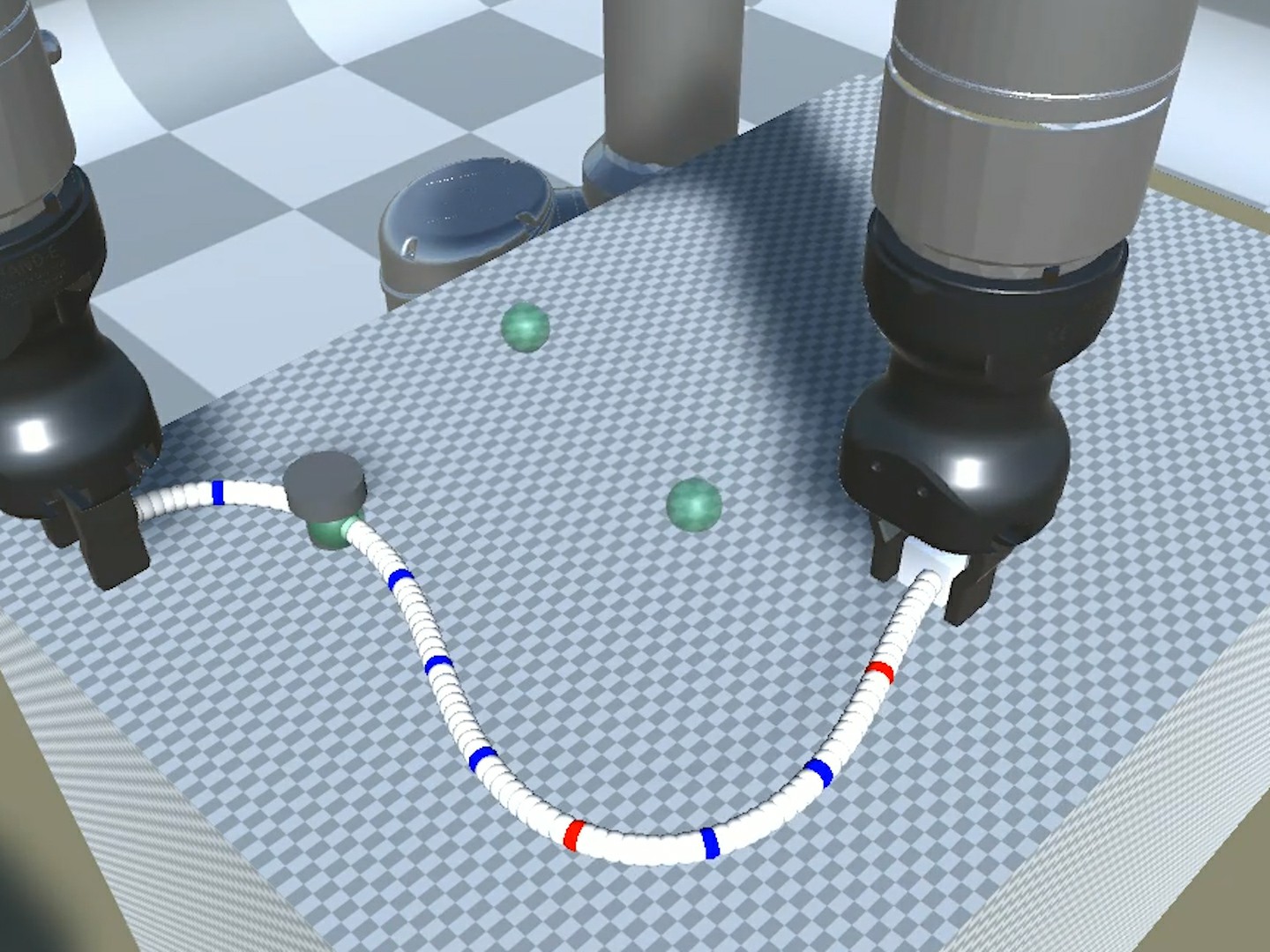} 
  }
  \hspace{-0.4cm}
  \subfigure[]{ 
    \includegraphics[width=3.4cm]{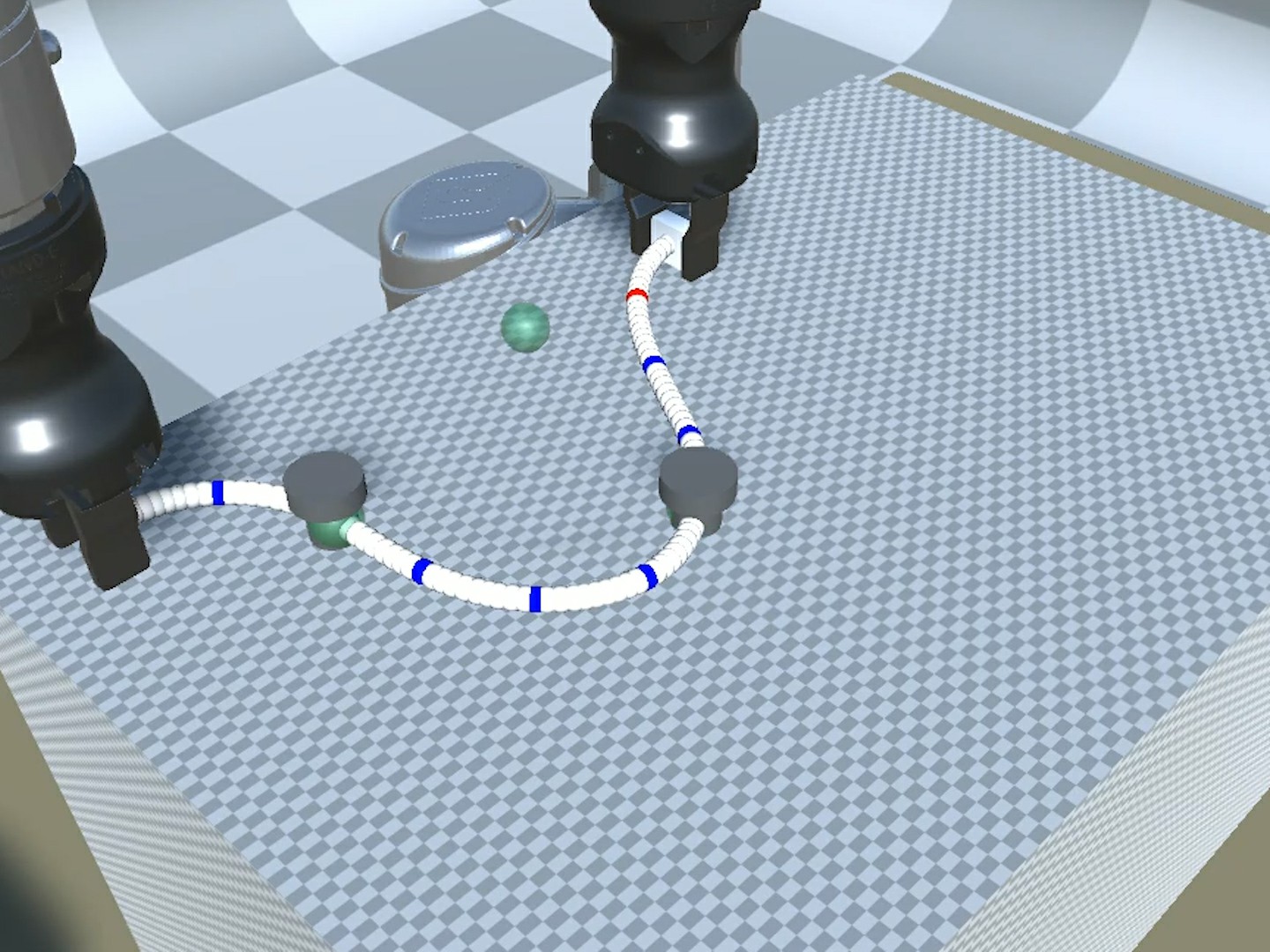} 
  } 
  \hspace{-0.4cm}
  \subfigure[]{ 
    \includegraphics[width=3.4cm]{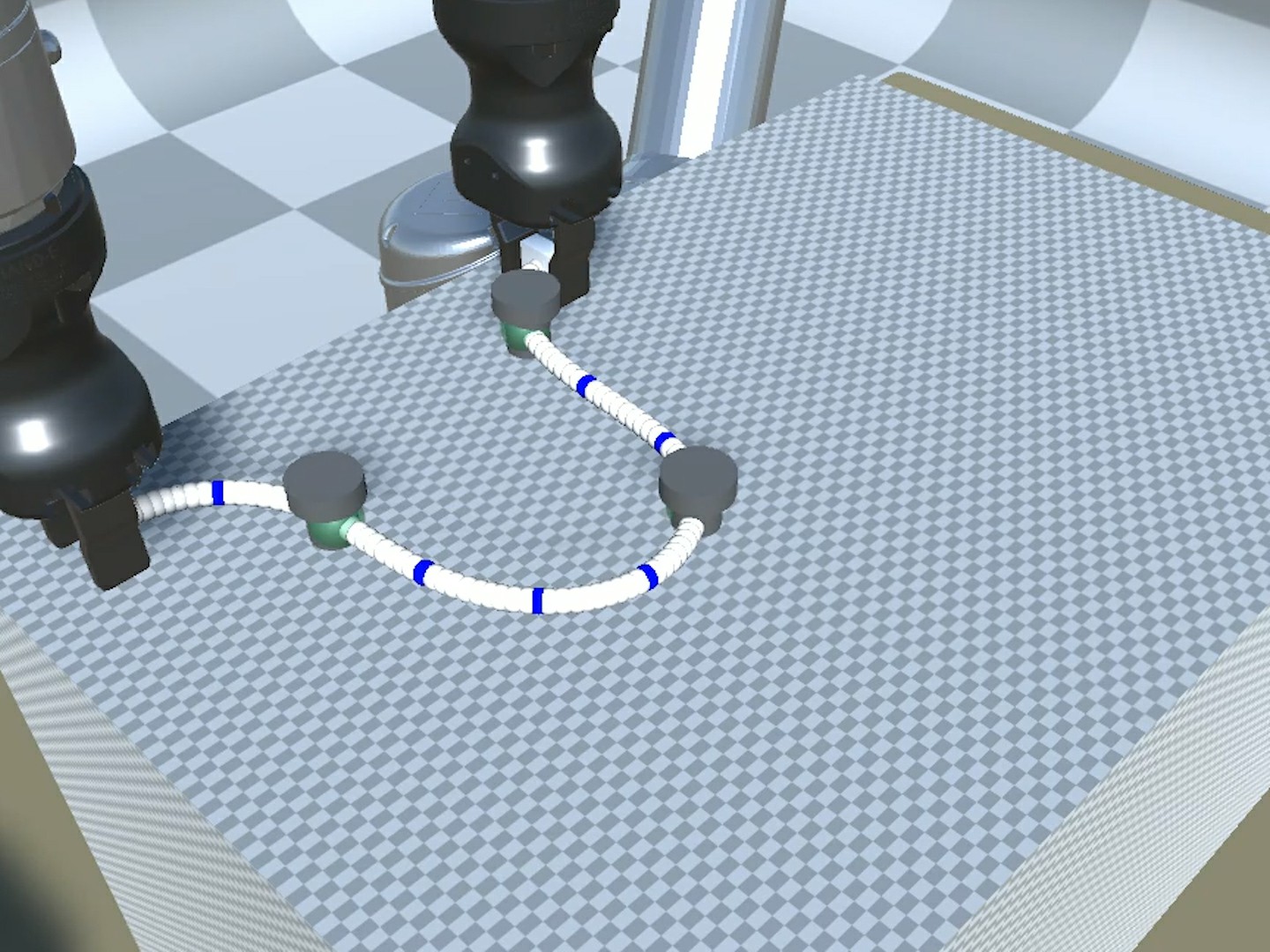} 
  } 
  \caption{Snapshots of the manipulation task 3. The DLO was manipulated to a shape like letter "U". Three target points on the DLO were manipulated sequentially. 
  (a) $t = 0 \hspace{0.2mm} {\rm s}$: The initial state. 
  (b) $t = 10 \hspace{0.2mm} {\rm s}$: The $1^{\rm st}$ target point reached its desired position. 
  (c) $t = 14 \hspace{0.2mm} {\rm s}$: The $1^{\rm st}$ target point was fixed by a nail then the $2^{\rm nd}$ target point was manipulated. 
  (d) $t = 32 \hspace{0.2mm} {\rm s}$: The $2^{\rm nd}$ target point reached its desired position and was fixed by a nail. 
  (e) $t = 53 \hspace{0.2mm} {\rm s}$: The $3^{\rm rd}$ target point reached its desired position and was fixed by a nail. The task was completed.}
  \label{fig:exp3}
\end{figure*}

\begin{figure} [!tb]
  \vspace{0.2cm}
  \centering 
  \subfigure[]{ 
    \includegraphics[width=8.5cm]{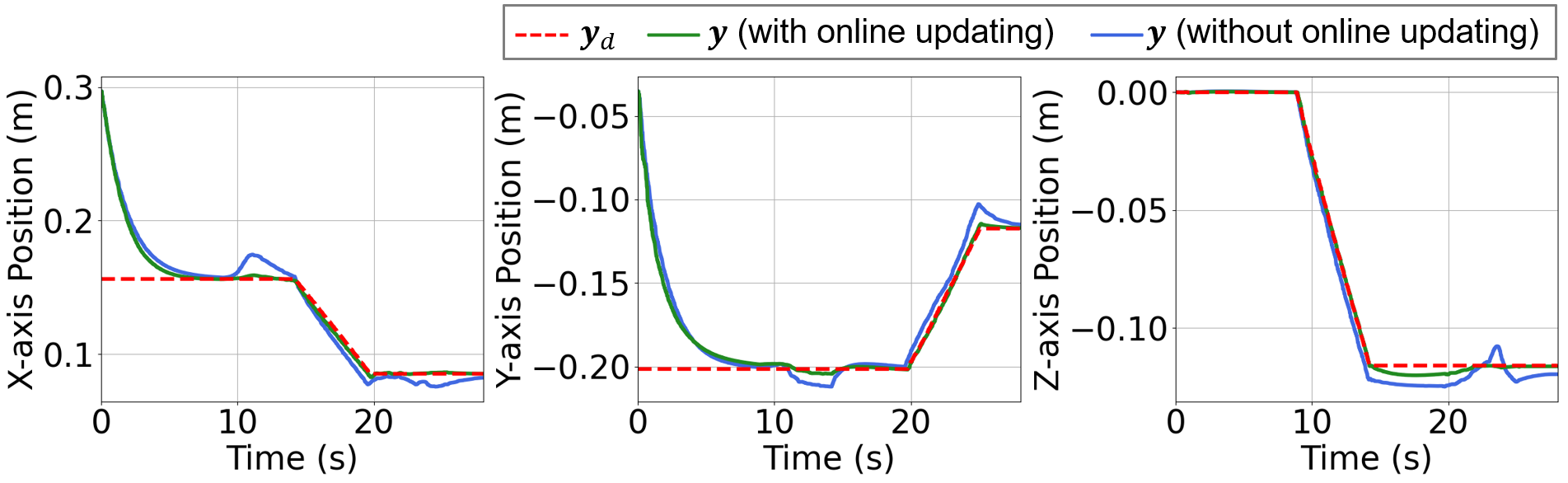} 
  } 
  \subfigure[]{ 
    \includegraphics[width=7.5cm]{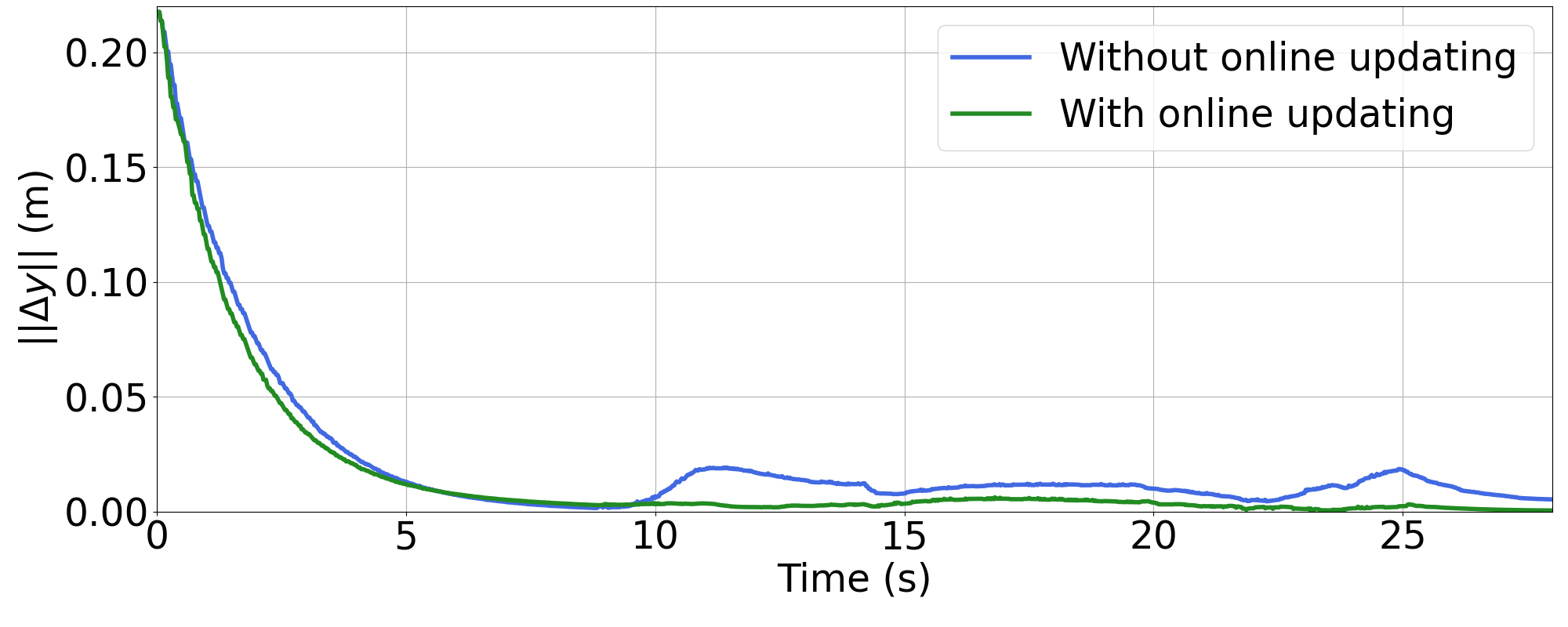} 
  }
  \vspace{-0.1cm}
  \caption{The results of task 2. 
  (a) The desired time-varying path and the actual path.
  (b) The position error between the desired path and the actual path during the manipulation task.} 
  \label{fig:exp2_deltaY}
    \vspace{-0.3cm}
\end{figure}

In the phase of the manipulation, the robot end effector was controlled to move the target point on the DLO to the desired position. The robot was referring to the trained NN in the offline phase and updating it again in parallel to the manipulation task. Three manipulation tasks were designed to test the performance of the proposed adaptive control approach with online learning.

The first manipulation task is shown in Fig. \ref{fig:exp1}. In this scenario, the target point was set as the fifth feature, which was manipulated to a fixed desired position. The NN trained with only 5-minute data in the offline phase was used. To show the effect of the online updating, the manipulation task was repeated twice, i.e. with or without the online updating. Fig. \ref{fig:online_origin} shows the comparison of these two manipulation processes, where both of them achieved the task but the online updating of the NN enabled the target point to be manipulated to the desired position faster, since the NN was updated to better adapt to the specific task using the updating law (\ref{control_update}). The parameters in (\ref{control_law}) and (\ref{control_update}) were set as $\bm K_p = {\rm diag}(0.2)$, $\bm L_i = {\rm diag}(20.0)$, $\lambda = 10.0$.

In the second manipulation task, 
the target point was controlled to follow the desired path, which was planned manually beforehand. 
The manipulation task is shown in Fig. \ref{fig:exp2cylinder}, in which the DLO was manipulated to encircle the cylinder. The target point was set as the sixth feature. The NN trained with 5-minute data in the offline phase was used. The manipulation task was also repeated twice, i.e. with or without the online updating. The comparison is shown in Fig. \ref{fig:exp2_deltaY}, where the position error of the manipulation with online updating was smaller. This was mainly because the online updating guaranteed the smaller model estimation error and hence led to the smaller manipulation error from (\ref{control_loop_equation_4}). The parameters in (\ref{control_law}) and (\ref{control_update}) were set as $\bm K_p = {\rm diag}(0.5)$, $\bm L_i = {\rm diag}(20.0)$, $\lambda = 10.0$. The results prove that the proposed control scheme can be used to achieve the relatively complicated manipulation task with a single target point under the proper planning. 


\begin{figure} [!tb]
  \centering 
  \subfigure[]{ 
    \includegraphics[width=2.8cm]{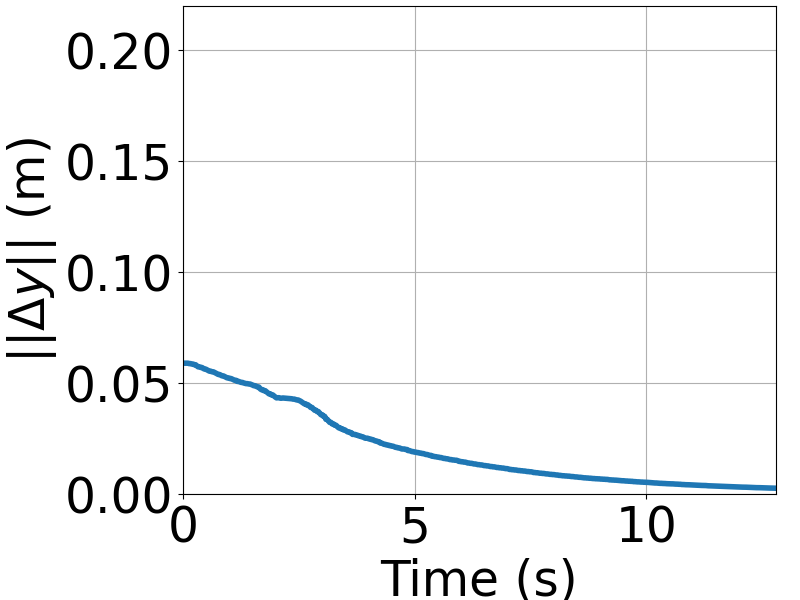} 
  } 
  \hspace{-0.5cm}
  \subfigure[]{ 
    \includegraphics[width=2.8cm]{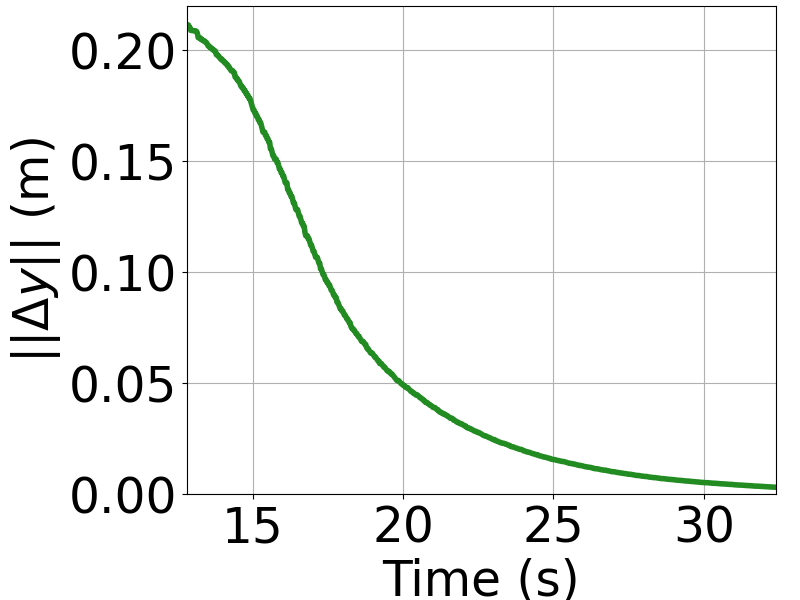}  
  }
  \hspace{-0.5cm}
  \subfigure[]{ 
    \includegraphics[width=2.8cm]{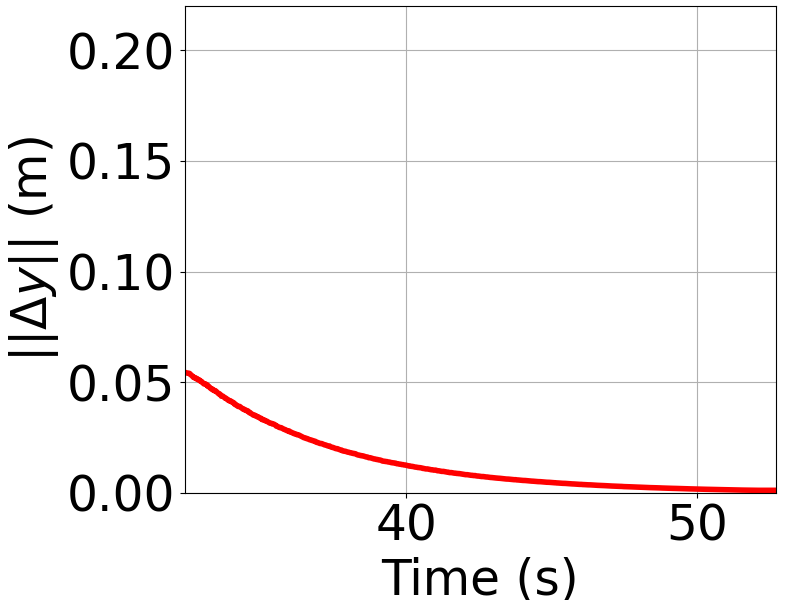}  
  }
  \vspace{-0.2cm}
  \caption{The results of task 3. Multiple target points were manipulated sequentially,, and $\| \Delta \bm y \|$ is the distance between the actual position and the desired position of the controlled target point.
  (a) The $1^{\rm st}$ target point.
  (b) The $2^{\rm nd}$ target point.
  (c) The $3^{\rm rd}$ target point.
  } 
  \label{fig:exp3_deltaY}
\vspace{-0.3cm}
\end{figure}

In the third manipulation task, multiple target points along the DLO were manipulated sequentially. Each target point was assigned a desired position. When the first target point was located at the desired position, it was fixed by external forces (e.g. hammering a nail on it). Then, the next target point was activated, which was also manipulated then fixed at its desired position. Such process was repeated until all the target points were fixed at their desired positions. 
Noted that the overall length of the DLO would change when the previous target points were fixed. Thus, the deformation model of the DLO was changing, which made the task challenging. The manipulation task is shown in Fig. \ref{fig:exp3}. The robot was controlled to manipulate the DLO to a overall shape like letter ``U''. Since the DLO was on a table, the vertical direction of the control input $\bm u$ was manually set as $0$. Other settings were all the same as those in the previous tasks. The three target points were set as the second, sixth and ninth features. The NN trained with 60-minute data in the offline phase was used. Fig. \ref{fig:exp3_deltaY} shows the manipulation error during the task. The parameters in (\ref{control_law}) and (\ref{control_update}) were set as $\bm K_p = {\rm diag}(0.2)$, $\bm L_i = {\rm diag}(1.0)$, $\lambda = 10.0$.


\hspace{-0.35cm}\textbf{Remark}:
The singular configurations of the estimated Jacobian matrix $\hat{\bm J}(\bm{\phi})$ can be found by carrying out the singular value decomposition, i.e. $\hat{\bm J}(\bm{\phi}) = \bm{U}\bm{\Sigma}\bm{V}^T = \sum_{i=1}^{l}{{\sigma_i} \bm u_i\bm v_i^T}$.
Hence,
a small $\sigma_i$ will result in the singularity and may lead to a large control input from (\ref{control_law}). In actual implementations, those terms with small $\sigma_i$ can be simply ignored in the summation to deal with the singular issues. 

\section{Conclusions}

This paper considers the robotic manipulation of DLOs with unknown deformation model, where the unknown model has been estimated in both the offline and the online phases. Both phases complement each other. That is, the offline learning can well initiate the estimation for the manipulation task, and the online learning can further reduce the approximation errors during the manipulation. 
The adaptive control scheme is proposed to achieve the manipulation task in the presence of the unknown deformation model.
The convergence of the task errors has been rigorously proved with Lyapunov methods, and simulation results in different scenarios have been presented.
Future works will be devoted to the validation of the proposed method on a real robot.

\addtolength{\textheight}{-12cm}   









\begin{thebibliography}{10}
\providecommand{\url}[1]{#1}
\csname url@rmstyle\endcsname
\providecommand{\newblock}{\relax}
\providecommand{\bibinfo}[2]{#2}
\providecommand\BIBentrySTDinterwordspacing{\spaceskip=0pt\relax}
\providecommand\BIBentryALTinterwordstretchfactor{4}
\providecommand\BIBentryALTinterwordspacing{\spaceskip=\fontdimen2\font plus
\BIBentryALTinterwordstretchfactor\fontdimen3\font minus
  \fontdimen4\font\relax}
\providecommand\BIBforeignlanguage[2]{{%
\expandafter\ifx\csname l@#1\endcsname\relax
\typeout{** WARNING: IEEEtran.bst: No hyphenation pattern has been}%
\typeout{** loaded for the language `#1'. Using the pattern for}%
\typeout{** the default language instead.}%
\else
\language=\csname l@#1\endcsname
\fi
#2}}

\bibitem{XEROULIS2007442}
G.~J. Xeroulis, J.~Park, C.-A. Moulton, R.~K. Reznick, V.~LeBlanc, and
  A.~Dubrowski, ``Teaching suturing and knot-tying skills to medical students:
  A randomized controlled study comparing computer-based video instruction and
  (concurrent and summary) expert feedback,'' \emph{Surgery}, vol. 141, no.~4,
  pp. 442--449, 2007.

\bibitem{8994188}
L.~{Cao}, X.~{Li}, P.~T. {Phan}, A.~M.~H. {Tiong}, H.~L. {Kaan}, J.~{Liu},
  W.~{Lai}, Y.~{Huang}, H.~M. {Le}, M.~{Miyasaka}, K.~Y. {Ho}, P.~W.~Y. {Chiu},
  and S.~J. {Phee}, ``Sewing up the wounds: A robotic suturing system for
  flexible endoscopy,'' \emph{IEEE Robotics Automation Magazine}, vol.~27,
  no.~3, pp. 45--54, 2020.

\bibitem{5642168}
A.~{Loeve}, P.~{Breedveld}, and J.~{Dankelman}, ``Scopes too flexible...and too
  stiff,'' \emph{IEEE Pulse}, vol.~1, no.~3, pp. 26--41, 2010.

\bibitem{8460694}
X.~{Li}, X.~{Su}, Y.~{Gao}, and Y.~{Liu}, ``Vision-based robotic grasping and
  manipulation of usb wires,'' in \emph{2018 IEEE International Conference on
  Robotics and Automation (ICRA)}, 2018, pp. 3482--3487.

\bibitem{8698220}
F.~{Zhong}, Y.~{Wang}, Z.~{Wang}, and Y.~{Liu}, ``Dual-arm robotic needle
  insertion with active tissue deformation for autonomous suturing,''
  \emph{IEEE Robotics and Automation Letters}, vol.~4, no.~3, pp. 2669--2676,
  2019.

\bibitem{7139532}
{Weifu Wang}, D.~{Berenson}, and D.~{Balkcom}, ``An online method for
  tight-tolerance insertion tasks for string and rope,'' in \emph{2015 IEEE
  International Conference on Robotics and Automation (ICRA)}, 2015, pp.
  2488--2495.

\bibitem{8403315}
T.~{Tang}, C.~{Wang}, and M.~{Tomizuka}, ``A framework for manipulating
  deformable linear objects by coherent point drift,'' \emph{IEEE Robotics and
  Automation Letters}, vol.~3, no.~4, pp. 3426--3433, 2018.

\bibitem{robotic_knitting_2020}
M.~Jovanovi{\'{c}}, M.~Vu{\v{c}}i{\'{c}}, B.~Tepav{\v{c}}evi{\'{c}},
  M.~Rakovi{\'{c}}, and J.~Tasevski, ``Robotic knitting in string art as a tool
  for creative design processes,'' in \emph{Advances in Service and Industrial
  Robotics}, K.~Berns and D.~G{\"o}rges, Eds.\hskip 1em plus 0.5em minus
  0.4em\relax Cham: Springer International Publishing, 2020, pp. 179--187.

\bibitem{JIMENEZ2012154}
P.~Jiménez, ``Survey on model-based manipulation planning of deformable
  objects,'' \emph{Robotics and Computer-Integrated Manufacturing}, vol.~28,
  no.~2, pp. 154--163, 2012.

\bibitem{ijrr2018}
J.~Sanchez, J.-A. Corrales, B.-C. Bouzgarrou, and Y.~Mezouar, ``Robotic
  manipulation and sensing of deformable objects in domestic and industrial
  applications: a survey,'' \emph{The International Journal of Robotics
  Research}, vol.~37, no.~7, pp. 688--716, 2018.

\bibitem{5509462}
M.~{Higashimori}, K.~{Yoshimoto}, and M.~{Kaneko}, ``Active shaping of an
  unknown rheological object based on deformation decomposition into elasticity
  and plasticity,'' in \emph{2010 IEEE International Conference on Robotics and
  Automation}, 2010, pp. 5120--5126.

\bibitem{Huan_ijrr_2015}
H.~Lin, F.~Guo, F.~Wang, and Y.-B. Jia, ``Picking up a soft 3d object by
  “feeling” the grip,'' \emph{The International Journal of Robotics
  Research}, vol.~34, no.~11, pp. 1361--1384, 2015.

\bibitem{6327684}
T.~{Bretl} and Z.~{McCarthy}, ``Mechanics and quasi-static manipulation of
  planar elastic kinematic chains,'' \emph{IEEE Transactions on Robotics},
  vol.~29, no.~1, pp. 1--14, 2013.

\bibitem{Timothy_ijrr_2014}
T.~Bretl and Z.~McCarthy, ``Quasi-static manipulation of a kirchhoff elastic
  rod based on a geometric analysis of equilibrium configurations,'' \emph{The
  International Journal of Robotics Research}, vol.~33, no.~1, pp. 48--68,
  2014.

\bibitem{roussel_deformable_2014}
O.~Roussel and M.~Taïx, ``\BIBforeignlanguage{en}{Deformable linear object
  manipulation planning with contacts},'' in \emph{\BIBforeignlanguage{en}{Full
  day workshop at {IEEE}/{RSJ} {International} {Conference} on {Intelligent}
  {Robots} and {Systems} ({IROS})}}, 2014, p.~6.

\bibitem{alvarez_interactive_2016}
N.~Alvarez and K.~Yamazaki, ``\BIBforeignlanguage{en}{An interactive simulator
  for deformable linear objects manipulation planning},'' in
  \emph{\BIBforeignlanguage{en}{2016 {IEEE} {International} {Conference} on
  {Simulation}, {Modeling}, and {Programming} for {Autonomous} {Robots}
  ({SIMPAR})}}, 2016, pp. 259--267.

\bibitem{8256194}
H.~{Han}, G.~{Paul}, and T.~{Matsubara}, ``Model-based reinforcement learning
  approach for deformable linear object manipulation,'' in \emph{2017 13th IEEE
  Conference on Automation Science and Engineering (CASE)}, 2017, pp. 750--755.

\bibitem{yan_self-supervised_2020}
M.~{Yan}, Y.~{Zhu}, N.~{Jin}, and J.~{Bohg}, ``Self-supervised learning of
  state estimation for manipulating deformable linear objects,'' \emph{IEEE
  Robotics and Automation Letters}, vol.~5, no.~2, pp. 2372--2379, 2020.

\bibitem{zhu_dual-arm_2018}
J.~{Zhu}, B.~{Navarro}, P.~{Fraisse}, A.~{Crosnier}, and A.~{Cherubini},
  ``Dual-arm robotic manipulation of flexible cables,'' in \emph{2018 IEEE/RSJ
  International Conference on Intelligent Robots and Systems (IROS)}, 2018, pp.
  479--484.

\bibitem{lagneau_automatic_2020}
R.~{Lagneau}, A.~{Krupa}, and M.~{Marchal}, ``Automatic shape control of
  deformable wires based on model-free visual servoing,'' \emph{IEEE Robotics
  and Automation Letters}, vol.~5, no.~4, pp. 5252--5259, 2020.

\bibitem{navarro2016Automatic}
D.~{Navarro-Alarcon}, H.~M. {Yip}, Z.~{Wang}, Y.~{Liu}, F.~{Zhong}, T.~{Zhang},
  and P.~{Li}, ``Automatic 3-d manipulation of soft objects by robotic arms
  with an adaptive deformation model,'' \emph{IEEE Transactions on Robotics},
  vol.~32, no.~2, pp. 429--441, 2016.

\bibitem{navarro2018fourier}
D.~{Navarro-Alarcon} and Y.~{Liu}, ``Fourier-based shape servoing: A new
  feedback method to actively deform soft objects into desired 2-d image
  contours,'' \emph{IEEE Transactions on Robotics}, vol.~34, no.~1, pp.
  272--279, 2018.

\bibitem{zhu2019_3dDeformable}
Z.~{Hu}, T.~{Han}, P.~{Sun}, J.~{Pan}, and D.~{Manocha}, ``3-d deformable
  object manipulation using deep neural networks,'' \emph{IEEE Robotics and
  Automation Letters}, vol.~4, no.~4, pp. 4255--4261, 2019.

\bibitem{ogden1997non}
R.~W. Ogden, \emph{Non-linear elastic deformations}.\hskip 1em plus 0.5em minus
  0.4em\relax Courier Corporation, 1997.

\bibitem{henrich2012robot}
D.~Henrich and H.~W{\"o}rn, \emph{Robot manipulation of deformable
  objects}.\hskip 1em plus 0.5em minus 0.4em\relax Springer Science \& Business
  Media, 2012.

\bibitem{sadd2009elasticity}
M.~H. Sadd, \emph{Elasticity: theory, applications, and numerics}.\hskip 1em
  plus 0.5em minus 0.4em\relax Academic Press, 2009.

\bibitem{xli2014adpative}
X.~{Li} and C.~C. {Cheah}, ``Adaptive neural network control of robot based on
  a unified objective bound,'' \emph{IEEE Transactions on Control Systems
  Technology}, vol.~22, no.~3, pp. 1032--1043, 2014.

\bibitem{kingma2014adam}
D.~P. Kingma and J.~Ba, ``Adam: A method for stochastic optimization,''
  \emph{arXiv preprint arXiv:1412.6980}, 2014.

\bibitem{arimoto1996control}
S.~Arimoto, \emph{Control theory of nonlinear mechanical systems}.\hskip 1em
  plus 0.5em minus 0.4em\relax Oxford University Press, 1996.

\bibitem{unity}
\BIBentryALTinterwordspacing
U.~Technologies. (2021) Unity real-time development platform. [Online].
  Available: \url{https://unity.com/}
\BIBentrySTDinterwordspacing

\bibitem{obi}
\BIBentryALTinterwordspacing
V.~M. Studio. (2019) Obi - {Unified} particle physics for {Unity 3D}. [Online].
  Available: \url{http://obi.virtualmethodstudio.com/}
\BIBentrySTDinterwordspacing

\bibitem{quigley2009ros}
M.~Quigley, K.~Conley, B.~Gerkey, J.~Faust, T.~Foote, J.~Leibs, R.~Wheeler,
  A.~Y. Ng, \emph{et~al.}, ``{ROS}: an open-source robot operating system,'' in
  \emph{ICRA workshop on open source software}, vol.~3, no. 3.2.\hskip 1em plus
  0.5em minus 0.4em\relax Kobe, Japan, 2009, p.~5.

\bibitem{RosSharp}
\BIBentryALTinterwordspacing
M.~Bischoff. (2021) {ROS\#}. [Online]. Available:
  \url{https://github.com/siemens/ros-sharp}
\BIBentrySTDinterwordspacing

\bibitem{pytorch}
A.~Paszke, S.~Gross, F.~Massa, A.~Lerer, J.~Bradbury, G.~Chanan, T.~Killeen,
  Z.~Lin, N.~Gimelshein, L.~Antiga, A.~Desmaison, A.~Kopf, E.~Yang, Z.~DeVito,
  M.~Raison, A.~Tejani, S.~Chilamkurthy, B.~Steiner, L.~Fang, J.~Bai, and
  S.~Chintala, ``Pytorch: An imperative style, high-performance deep learning
  library,'' in \emph{Advances in Neural Information Processing Systems
  32}.\hskip 1em plus 0.5em minus 0.4em\relax Curran Associates, Inc., 2019,
  pp. 8024--8035.

\end{thebibliography}

\bibliographystyle{IEEEtran}

\end{document}